\documentclass{article}

\usepackage{microtype}
\usepackage{graphicx}
\usepackage{subfigure}
\usepackage{booktabs} % for professional tables

\usepackage{hyperref}

\usepackage[accepted]{icml2018_arxiv}

\usepackage{amsmath}
\usepackage{amssymb}
\usepackage{amsthm}
\usepackage{bm}
\usepackage{times}
\usepackage{graphicx} % more modern
\usepackage{subfigure} 
\usepackage{multirow}
\usepackage{color}
\usepackage{algorithm}
\usepackage{algorithmic}
\usepackage{multirow}

\newtheorem{theorem}{Theorem}[section]

\newtheorem{proposition}[theorem]{Proposition}

\icmltitlerunning{Learning Registered Point Processes from Idiosyncratic Observations}

\begin{document}

\twocolumn[
\icmltitle{Learning Registered Point Processes from Idiosyncratic Observations}



\icmlsetsymbol{equal}{*}

\begin{icmlauthorlist}
\icmlauthor{Hongteng Xu}{du,inf}
\icmlauthor{Lawrence Carin}{du}
\icmlauthor{Hongyuan Zha}{gt}
\end{icmlauthorlist}

\icmlaffiliation{du}{Department of ECE, Duke University, Durham, NC, USA}
\icmlaffiliation{inf}{InfiniaML Inc., Durham, NC, USA}
\icmlaffiliation{gt}{College of Computing, Georgia Institute of Technology, Atlanta, GA, USA}

\icmlcorrespondingauthor{Hongteng Xu}{hongteng.xu@duke.edu}
%\icmlcorrespondingauthor{Eee Pppp}{ep@eden.co.uk}

\icmlkeywords{Registered point processes, idiosyncratic observations, warping functions, maximum likelihood estimation}

\vskip 0.3in
]



\printAffiliationsAndNotice{}  % leave blank if no need to mention equal contribution

\begin{abstract}
A parametric point process model is developed, with modeling based on the assumption that sequential observations often share latent phenomena, while also possessing idiosyncratic effects. 
An alternating optimization method is proposed to learn a ``registered'' point process that accounts for shared structure, as well as ``warping'' functions that characterize idiosyncratic aspects of each observed sequence. 
Under reasonable constraints, in each iteration we update the sample-specific warping functions by solving a set of constrained nonlinear programming problems in parallel, and update the model by maximum likelihood estimation. 
The justifiability, complexity and robustness of the proposed method are investigated in detail, and the influence of sequence stitching on the learning results is examined empirically.
Experiments on both synthetic and real-world data demonstrate that the method yields explainable point process models, achieving encouraging results compared to state-of-the-art methods.
\end{abstract}

\section{Introduction}
The behavior of real-world entities often may be recorded as event sequences; for example, interactions of participants in a social network, the admissions of patients, and the job-hopping behavior of employees. 
In practice, these behaviors are under the control of complicated mechanisms, which can be captured approximately by an appropriate parametric temporal point process model. 
While the observed event sequences associated with a given process ($e.g.$, disease) may share common (``standard'') attributes, there are often subject-specific factors that may impact the observed data. 
For example, the admission records of different patients are always personalized: even if the patients suffer from the same disease, they may spend unequal time on recovery because their medications, history and environmental conditions may be distinct. 
Another typical example is the job-hopping behavior of employees. 
The employees in the same company often make very different career plans depending on their age, family situation and unobserved status of the job market. 

\begin{figure}[t]
\centering
\includegraphics[width=0.9\linewidth]{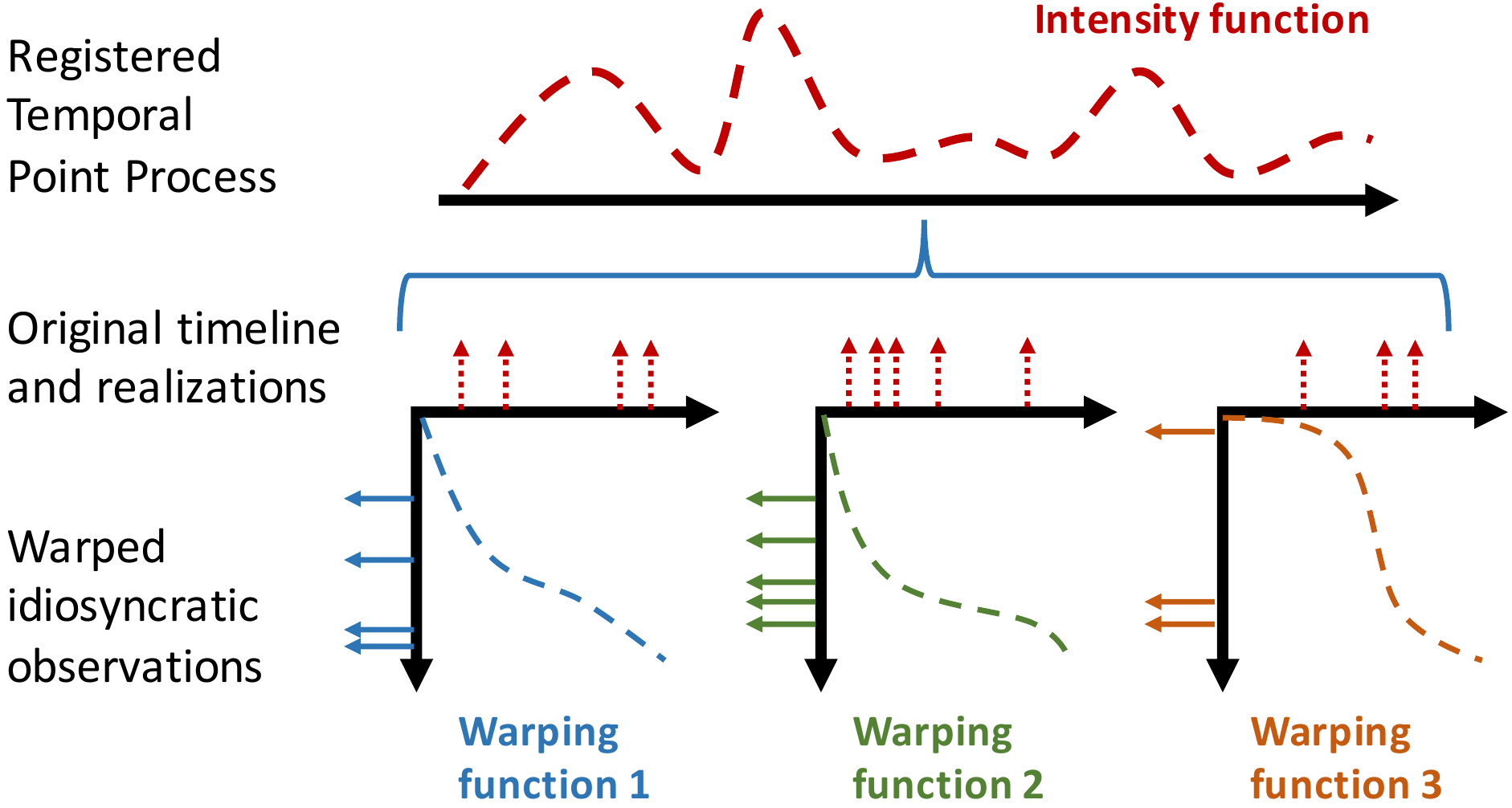}\vspace{-10pt}
\caption{\small The illustration of concepts in our work. The dotted parts (parametric point process model, unwarped realizations and warping functions) are what we aim to learn.\label{illu}}
\end{figure}

The examples above reveal that event sequences that share an underlying temporal point process, linked to a given phenomenon of interest, may be personalized by hidden idiosyncratic factors; this may be represented as a subject-specific ``warping'' along time, as shown in Fig.~\ref{illu}. 
The characteristics of such data often have a negative influence on the learning of the target point process, $i.e.$, increase the uncertainty of the model. 
The complexity of models can be increased to fit the personalized observations well, $e.g.$, the locally-stationary point processes in~\cite{roueff2016locally,mammen2017nonparametric,xu17b}. 
However, from the viewpoint of model registration, it is desirable to separate the essential mechanism of the model and idiosyncratic aspects of the data, such that the final model is ``registered'' and characterizes the shared phenomena, while also inferring what is sample-specific.

Learning registered point processes from idiosyncratic observations is a challenging problem, requiring one to jointly learn a shared point process model and a set of sample-specific warping functions. 
To solve this problem, we propose a novel and effective learning method based on alternating optimization. 
Specifically, in each iteration we first apply the inverse of estimated warping functions ($i.e.$, unwarping functions) to unwarp observed event sequences and learn the parameter of a registered point process by maximum likelihood estimation; we then update the warping functions of event sequences, based on the estimation of registered point process.
The new functions are applied to update the model for the next iteration.
In particular, we approximate the warping/unwarping functions of event sequences by piecewise linear models, and learn their parameters by solving a set of constrained nonlinear programming problems in parallel. 

We analyze the justification for and the complexity of our method in detail.
The meaning of the regularizers and constraints used in our method, and their effects, are investigated. 
Further, we consider improved learning by stitching warped sequences randomly and learning from the stitched sequences, verifying the feasibility of this data processing strategy empirically. 
Experimental results show that the proposed method outperforms its alternatives on both synthetic and real-world data.  

\section{Proposed Model}
Denote a parametric temporal point process as $N_{\theta}$. 
Its event sequence consists of multiple events $\{(t_i, c_i)\}_{i=1}^{I}$ with time stamps $t_i \in [0,T]$ and event types $c_i\in \mathcal{C}=\{1,...,C\}$, which can be represented as $\{N_c(t)\}_{c=1}^{C}$, where $N_c(t)$ is the number of type-$c$ events occurring at or before time $t$. 
A temporal point process can be characterized by its intensity functions $\{\lambda_c(t)\}_{c=1}^C$, where $\lambda_c(t)={\mathbb{E}[dN_c(t)|\mathcal{H}_t^{\mathcal{C}}]}/{dt}$ and $\mathcal{H}_t^{\mathcal{C}}= \{(t_i, c_i) |  t_i < t, c_i\in\mathcal{C}\}$ collects historical events before time $t$. 
Each $\lambda_c(t)$ represents the expected instantaneous rate of the type-$c$ event at time $t$, which can be parametrized by $\theta$. 

We assume ($i$) an exponential-like intensity: each $\lambda(t)$ is represented $\sum_{j=1}^{J}\exp_{t_j}(f_j(t;\theta,\mathcal{H}_t^{\mathcal{C}}))$, where $J\geq 1$, the $f_j$ are linear functions of time, which are related to $\theta$ and historical observations; 
$\exp_{t_j}(f_j(t))=\exp(f_j(t))$ if $t\geq t_j$, otherwise, it equals to $0$. 
Note that many important point processes, $e.g.$, the Hawkes process~\cite{hawkes1974cluster} and the self-correcting process~\cite{isham1979self,xu2015trailer,xu2017patient} satisfy this assumption (see Appendix~\ref{ap1}). 

The sequences of $N_{\theta}$ may be warped in $[0,T]$ by a set of continuous and invertible warping functions. 
Denote the sequences and the corresponding warping functions as $\{S_m\}_{m=1}^M$ and $\{W_m\}_{m=1}^{M}$, respectively. 
Each $S_m=\{(t_i^m, c_i^m)\}_{i=1}^{I_m}$ contains $I_m$ events, whose time stamps are deformed from a ``standard'' timeline under the corresponding warping function $W_m:~[0,T]\mapsto [0,T]$. 
Accordingly, the unwarping functions can be denoted as $\{W_m^{-1}\}_{m=1}^{M}$.
For $m=1,...,M$, we assume\footnote{Different from~\cite{tang2008pairwise,panaretos2016amplitude}, which imposes these two assumptions on warping functions, we impose them on unwarping functions, to for simplify the following learning algorithm. 
In such a situation, the warping functions may disobey the unbiasedness assumption. 
Fortunately, when the unwarping functions satisfy these two assumptions, $\mathbb{E}[W_m(t)]$ is can still be close to an identity function in most common cases.}
($ii$) unbiasedness: $\mathbb{E}[W_m^{-1}(t)] = t$ on $[0, T]$, and ($iii$)
regularity: $W_m^{-1}(t)$ is monotone increasing on $[0, T]$. 

Taking the warping functions into account, the likelihood of an (unobserved) unwarped sequence can be formulated based on the intensity functions~\cite{daley2007introduction}:
\begin{eqnarray}\label{mlew}
\begin{aligned}
\mathcal{L}(\theta;W_m^{-1}(S_m))=\frac{\prod_{i=1}^{I_m}\lambda_{c_i^m}(W_m^{-1}(t_i^m))}{\exp\left(\sum_{c=1}^{C}\int_{0}^{T}\lambda_c(W_m^{-1}(s))ds\right)},
\end{aligned}
\end{eqnarray}
where $W_m^{-1}(S_m)$ represents the unwarped event sequence, $i.e.$, $W_m^{-1}(S_m)=\{(W_m^{-1}(t_i^m), c_i^m)\}_{i=1}^{I_m}$. 

The warped data caused by idiosyncratic effects generally do harm to the maximum likelihood estimation of the target point process, except for some trivial cases (See Appendix~\ref{ap7}):
\begin{proposition}\label{thm1}
For a temporal point process $N_\theta$ satisfying Assumption ($i$) from above, $\hat{\theta}^*$ and $\hat{\theta}$ denote its maximum likelihood estimation based on original data and that based on warped data, respectively. 
Then $\hat{\theta}^*=\hat{\theta}$ if and only if 1) the warping functions are translations; or 2) $N_\theta$ is a homogeneous Poisson process.
\end{proposition}
The problem is that given the warped observations $\{S_m\}_{m=1}^M$, we seek to learn a ``registered'' model $\theta$, as well as sample-specific warping functions $\{W_m\}_{m=1}^{M}$ (or equivalently, the unwarping functions $\{W_m^{-1}\}_{m=1}^{M}$) . 

\section{Learning Registered Point Processes}
\subsection{Maximizing the likelihood}
We develop a learning method based on maximum likelihood estimation (MLE). 
Considering the assumptions of unwarping functions and the likelihood in (\ref{mlew}), we can formulate the optimization problem as
\begin{eqnarray}\label{opt}
\begin{aligned}
\min_{\theta,\{W_m\}}~&-\sideset{}{_{m}}\sum\log \mathcal{L}(\theta;W_m^{-1}(S_m))+\gamma \mathcal{R}(\{W_m^{-1}\})\\
s.t.~&1)~W_m^{-1}(0) = 0,~W_m^{-1}(T) = T,~\text{and}\\
&2)~{W_m^{-1}}'(t)>0~\text{for $m=1,...,M$}, 
\end{aligned}
\end{eqnarray}
where ${W_m^{-1}}'(t)=\frac{dW_m'}{dt}$ is the derivative of unwarping function. 
In our objective function, the first term represents the negative log-likelihood of unwarped event sequences while the second term represents the regularizer imposed on unwarping functions. 
For each unwarping function, the first constraint corresponds to its range and the second constraint makes it obey the regularity assumption.
Furthermore, according to the unbiasedness assumption, we apply the following regularizer:
\begin{eqnarray}\label{reg}
\begin{aligned}
\mathcal{R}(\{W_m^{-1}\}) = \int_{0}^{T}\Bigl|\frac{1}{M}\sideset{}{_{m=1}^{M}}\sum W_m^{-1}(s) - s\Bigr|^2 ds. 
\end{aligned}
\end{eqnarray}

The optimization problem in (\ref{opt}) is non-convex and has a large number of unknown variables. 
Solving it directly is intractable.
Fortunately, for the parametric point processes with exponential-like intensity functions, we can design an effective alternating optimization method to solve the problem iteratively, after parameterizing the warping functions as piecewise linear. 
In each iteration, we first maximize the likelihood of the unwarped sequences based on the estimation of warping functions, and then optimize the warping functions based on the estimated model. 

Specifically, in the $k$-th iteration, given the warping functions estimated in the previous iteration, $i.e.$, $\{W_m^{k-1}\}_{m=1}^{M}$, we learn the target point process by
\begin{eqnarray}\label{opt:mle}
\begin{aligned}
\theta^{k}=\arg\sideset{}{_{\theta}}\min-\sideset{}{_{m=1}^{M}}\sum\log \mathcal{L}(\theta;(W_m^{k-1})^{-1}(S_m)).
\end{aligned}
\end{eqnarray}
Focusing on different point processes, we can apply various optimization methods to solve this problem. 
For example, learning Hawkes processes can be achieved in the framework of expectation-maximization (EM)~\cite{lewis2011nonparametric,zhou2013learning}, which is equivalent to a projected-gradient-ascent algorithm. 
For other kinds of parametric point processes, $e.g.$, the self- and mutually-correcting processes, we can learn their parameters by gradient descent or stochastic gradient descent (SGD).

\subsection{Learning warping/unwarping functions}\label{ssec:warp}
Given $\theta^{k}$, seek to update the warping/unwarping functions. 
To simplify the problem and accelerate our learning method, we take advantage of the warping functions estimated in the previous iteration, $i.e.$, $\{W_m^{k-1}\}_{m=1}^{M}$, and decompose the problem into $M$ independent problems: for $m=1,...,M$, $W_m^k$ is the solution of
\begin{eqnarray}\label{opt:wm}
\begin{aligned}
&\sideset{}{_{W_m}}\min-\log \mathcal{L}(\theta^k; W_m^{-1}(S_m))\\
&+\gamma\int_{0}^{T}\Bigl|\frac{W_m^{-1}(s)}{M}+\frac{\sum_{m'\neq m}(W_{m'}^{k-1})^{-1}(s)}{M} - s\Bigr|^2 ds\\
&s.t.~W_m^{-1}(0) = 0,~W_m^{-1}(T) = T,~{W_m^{-1}}'(t)>0.
\end{aligned}
\end{eqnarray}
Solving these problems is non-trivial, requiring further parameterization of the warping functions $\{W_m\}_{m=1}^{M}$, or equivalently, the unwarping functions $\{W_m^{-1}\}_{m=1}^{M}$. 

We apply a set of piecewise linear models to fit the unwarping functions, for the convenience of mathematical derivation and computation. 
Specifically, given $L$ landmarks $\{t_1, ..., t_L\}$ in $[0, T]$, where $t_1=0$, $t_L=T$ and $t_l<t_{l+1}$, we model $W_m^{-1}$ for $m=1,...,M$ as
\begin{eqnarray}\label{fit}
\begin{aligned}
W_m^{-1}(t) = a_{l}^m t + b_{l}^m,~\text{if}~t\in [t_l, t_{l+1}).
\end{aligned}
\end{eqnarray}
Denoting $\bm{a}^m=\{a_l^m\}_{l=1}^{L-1}$ and $\bm{b}^{m}=\{b_l^m\}_{l=1}^{L-1}$ as the parameters of the model, we rewrite the regularizer and the constraints of $W_m^{-1}$ as
\begin{eqnarray}\label{cons1}
\begin{aligned}
&\int_{0}^{T}\Bigl|\frac{W_m^{-1}(s)}{M}+\frac{\sum_{m'\neq m}(W_{m'}^{k-1})^{-1}(s)}{M} - s\Bigr|^2 ds\\
&\rightarrow \Bigl\|\frac{1}{M}\bm{a}^{m}+\bm{a}^{\bar{m}}\Bigr\|_2^2 + \Bigl\|\frac{1}{M}\bm{b}^{m}+\bm{b}^{\bar{m}}\Bigr\|_2^2,\\
&W_m^{-1}(0) = 0\rightarrow b_{1}^m=0,\\
&W_m^{-1}(T) = T\rightarrow a_{L-1}^m T + b_{L-1}^m = T,\\
&{W_m^{-1}}'(t)>0\rightarrow a_{l}^m>0~\text{for}~l=1,...,L-1,
\end{aligned}
\end{eqnarray}
where $\|\cdot\|_2$ indicates the $\ell_2$ norm of a vector, $\bm{a}^{\bar{m}}=\frac{\sum_{m'\neq m}\bm{a}^{m',k-1}}{M}-\bm{1}$ and $\bm{b}^{\bar{m}}=\frac{\sum_{m'\neq m}\bm{b}^{m',k-1}}{M}$. 
$\bm{a}^{m',k-1}$ and $\bm{b}^{m',k-1}$ are estimated in the previous iteration. 
To guarantee continuity of $W_m^{-1}$, we further impose the following constraints on $\bm{a}^m$ and $\bm{b}^m$: for $l=1,...,L-2$, 
\begin{eqnarray}\label{cons2}
\begin{aligned}
a_{l}^m t_{l+1} + b_{l}^m = a_{l+1}^m t_{l+1} + b_{l+1}^m.
\end{aligned}
\end{eqnarray}

Based on the piecewise-linear model and the exponential-like intensity assumption, we propose a tight upper bound for the negative log-likelihood in (\ref{opt:wm}):
\begin{eqnarray}\label{surrogate}
\begin{aligned}
&-\log \mathcal{L}(\theta^k; W_m^{-1}(S_m))\\
=&\sum_{c=1}^{C}\int_{0}^{T}\lambda_c(W_m^{-1}(s))ds-\sum_{i=1}^{I_m}\log\lambda_{c_i^m}(W_m^{-1}(t_i^m))\\
\leq &\sideset{}{_{c=1}^{C}}\sum\int_{0}^{T}\lambda_c(s)dW_m(s)\\
&-\sideset{}{_{i=1}^{I_m}}\sum\sideset{}{_{j=1}^{J_i}}\sum q_{ij}^m\log({\lambda_{c_i^m}(W_m^{-1}(t_i^m))}/{q_{ij}^m})\\
=&\sum_{l=1}^{L-1}\Biggl[\frac{p_{l}^m}{a_{l}^m} - \sum_{j=1}^{J}\sum_{t_i^m\in[t_l, t_{l+1})}q_{ij}^m f_j(a_l^m t_i^m + b_l^m)\Biggr]+\mathsf{C}\\
=&\mathcal{Q}(\bm{a}^m,\bm{b}^m).
\end{aligned}
\end{eqnarray}
Here, $\lambda_{c_i^m}(W_m^{-1}(t_i^m))=\sum_{j=1}^{J}\exp(f_j(W_m^{-1}(t_i^m);\theta^{k}))$, the coefficients $p_{l}^m=\sum_c\int_{W_m^{-1}(t_{l})}^{W_m^{-1}(t_{l+1})}\lambda_c(s)ds$, $q_{ij}^m=\frac{\exp(f_j(W_m^{-1}(t_j^m)))}{\lambda_{c_i^m}(W_m^{-1}(t_i^m))}$ and $\mathsf{C}$ is the constant independent to $W_m^{-1}$. 
The inequality is based on Jensen's inequality and the $\{p_{l}^m, q_{ij}^m\}$ are calculated based on the parameters estimated in the previous iteration. 
The detailed derivation and the implementation for Hawkes process are given in Appendices~\ref{ap2} and~\ref{ap3}. 
Considering (\ref{cons1},~\ref{cons2},~\ref{surrogate}) together, we propose the surrogate problem of (\ref{opt:wm}):
\begin{eqnarray}\label{opt:ab}
\begin{aligned}
\min_{\bm{a}^m,\bm{b}^m}&\mathcal{Q}(\bm{a}^m,\bm{b}^m)+\gamma\Bigl\|\frac{\bm{a}^{m}}{M}+\bm{a}^{\bar{m}}\Bigr\|_2^2 + \gamma\Bigl\|\frac{\bm{b}^{m}}{M}+\bm{b}^{\bar{m}}\Bigr\|_2^2\\
s.t.~&1)~b_{1}^m=0,~a_{L-1}^m T + b_{L-1}^m = T,\\
&2)~\text{for}~l=1,...,L-1,~a_{l}^m>0,~\text{and}\\
&3)~a_{l}^m t_{l+1} + b_{l}^m = a_{l+1}^m t_{l+1} + b_{l+1}^m.
\end{aligned}
\end{eqnarray}
The setup in (\ref{opt:ab}) is a typical constrained nonlinear programming problem. 
Many optimization methods can be applied here, $e.g.$, sequential quadratic programming and an interior-point method. 
Note that estimating optimal $\bm{a}^m$ and $\bm{b}^m$, we need to re-calculate the $\{p_{l}^m, q_{ij}^m\}$ in $\mathcal{Q}$ and solve (\ref{opt:ab}) iteratively until convergence.

%\begin{algorithm}[t]
%\caption{\small Learning A Registered Point Process from Warped Observations (RPP)}\label{alg1}
%\textbf{Input:} Warped sequence $\{S_m\}_{m=1}^{M}$, weight of regularizer $\gamma$, the landmark points $\{t_l\}_{l=1}^{L}$, and the maximum numbers of outer and inner iterations.\\
%\textbf{Output:} Model's parameter $\theta$ and unwarping functions' parameters $\{\bm{a}^m,\bm{b}^m\}_{m=1}^{M}$.
%\begin{algorithmic}[1]
%   \STATE Set $\theta$ randomly, $\bm{a}^{m}=\bm{1}$, $\bm{b}^{m}=\bm{0}$ for $m=1,...,M$. 
%   \REPEAT
%   \STATE Given $\{\bm{a}^{m},\bm{b}^{m}\}_{m=1}^{M}$, unwarp data and update $\theta$ by solving (\ref{opt:mle}).
%   \FOR{$m=1,...,M$}
%   \REPEAT
%   \STATE Given $\{\bm{a}^{m},\bm{b}^{m},\theta\}$, calculate $\{p_{l}^m, q_{ij}^m\}$.
%   \STATE Update $\bm{a}^{m}$ and $\bm{b}^{m}$ by solving (\ref{opt:ab}).
%   \UNTIL{Converge or reach the maximum number of inner iterations}  
%   \ENDFOR 
%   \UNTIL{Converge or reach the maximum number of outer iterations}  
%\end{algorithmic}
%\end{algorithm}

%\subsection{Proposed scheme of algorithm}
Repeating the two steps above, we estimate the model and the warping/unwarping functions effectively.
%We denote our method as RPP and summarize its scheme in Algorithm~\ref{alg1}. 

\subsection{Justifiability Analysis}
The reasons for applying piecewise linear models to warping functions are twofold. 
First, our learning method involves computation of unwarping function $W_m^{-1}$ and the derivative of warping function $W_m^{\prime}$. 
Applying our piecewise linear model, both warping and unwarping functions can be represented explicitly. 
If we use other basis functions, $e.g.$, Gaussian basis, to represent $W_m$ (or $W_m^{-1}$), the $W_m^{-1}$ (or $W_m^{\prime}$) may be hard to be represented in closed-form. 
Second, compared to the finite element analysis used in functional optimization and differential equations, which discretizes functions into a grid, our piecewise linear model requires much fewer parameters, reducing the risk of over-fitting while also improving computational complexity.

\textbf{Complexity} 
Consider a $C$-dimensional Hawkes process as an example. 
We implement the MLE step and the updating of unwarping functions via an EM-based framework~\cite{zhou2013learning} and an interior-point method~\cite{potra2000interior}, respectively. 
Given $M$ sequences with $I$ events in each, the computational complexity of our method per iteration, in the worst case, is $\mathcal{O}(MI^2 + C^2 + ML^3)$. 
The $\mathcal{O}(MI^2)$ and $\mathcal{O}(C^2)$ correspond to the computational complexity of the E-step and the M-step, and the $\mathcal{O}(ML^3)$ corresponds to the computational complexity of solving $M$ nonlinear programming with $2L$ variables each, in the worst case. 
Because we update unwarping functions by solving $M$ independent optimization problems in parallel, the time complexity of our method can be $\mathcal{O}(MI^2 + C^2 + L^3)$. 

\textbf{Convergence} Our learning method converges in each step. 
For parametric point processes like Hawkes processes, their likelihood functions are convex and the convergence of the MLE-step is guaranteed. 
Further, the objective function in (\ref{opt:ab}) is convex, as shown in Appendix~\ref{ap4}, thus updating of the unwarping functions also converges well. 

Compared with existing methods, $e.g.$, the Wasserstein learning-based registration method (WLR)~\cite{bigot2012consistent,panaretos2016amplitude,zemel2017fr} and the multi-task learning-based method (MTL)~\cite{luo2015multi}, our RPP method has several advantages.
First, both WLR and the MTL require learning a specific model for each event sequence. 
For complicated multi-dimensional point processes, they require a large amount of events per sequence to learn reliable models independently, which might be unavailable in practice. 
Our method has much fewer parameters, and thus has much lower computational complexity and lower risk of over-fitting. 
Second, both the WLR and the MTL decompose learning of model and warping functions into two independent steps. 
The estimation error caused in the previous step will propagate to the following one.  
On the contrary, our method optimizes model and warping functions alternatively with guaranteed convergence, so the estimation error will be suppressed. 

\begin{figure}[t]
\centering
\includegraphics[width=0.4\columnwidth]{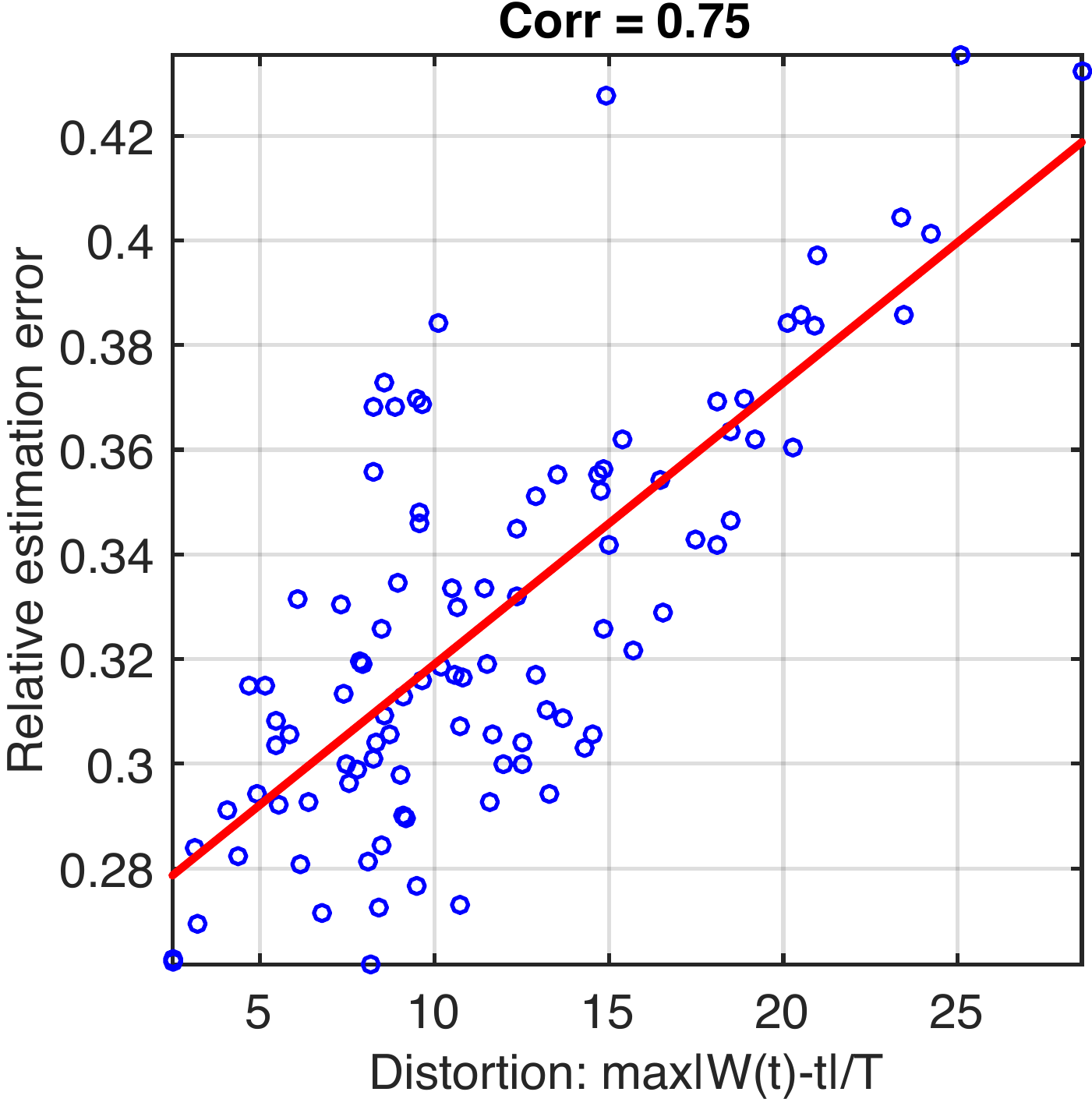}\label{distort}
\vspace{-10pt}
\caption{\small In $100$ trials, $40$ sequences with length $T$ are generated by a 1D Hawkes process and warped by a warping function with a certain $\|W(t)-t\|_{\infty}$. 
We learn the parameter of the model and record the points corresponding to the estimation errors and the proposed distortions, whose correlation is as high as $0.75$.}
\end{figure}

\section{Potential Improvement Based on Stitching}\label{sec:stitch}
Empirically, the influence of warped data on learning results is correlated with the distortion of warping function. 
The distortion should be a measurement not only dependent with the difference between warping function and identity function but also related to the scale of time window because the distortion on a certain scale becomes ignorable when we observe and analyze it on a larger scale with more samples.  
In particular, we propose a definition the distortion as $D=\frac{\|W(t)-t\|_{\infty}}{T}$.
Here, $\|W(t)-t\|_{\infty}=\max\{|W(t)-t|,~\forall t\in[0,T]\}$, which represents the most serious warping achieved by the warping function, and $T$ is the length of time window. 
In Fig.~\ref{distort}, we show that the distortion based on this definition is highly correlated with the relative estimation error ($i.e.$, $\frac{\|\theta^*-\theta\|_2}{\|\theta\|_2}$, where $\theta$ is the ground truth and $\theta^*$ is the estimation result). 

This relationship $\frac{\|\theta^*-\theta\|_2}{\|\theta\|_2} \propto D$ implies a potential strategy to further improving learning. 
Suppose that we have two warped sequences $S_1=\{(t_i^1, c_i^1)\}_{i=1}^{I_1}$ and $S_2=\{(t_i^2, c_i^2)\}_{i=1}^{I_2}$ observed in $[0,T]$, whose distortions are $D_1$ and $D_2$, respectively. 
If we stitch these two sequences together, $i.e.$, $S=S_1\cup\S_2=\{(t_1^1,c_1^1),...,(t_1^2+T, c_1^2),...\}$, the distortion of $S$ in $[0,2T]$ will be $D=\frac{1}{2}\max\{D_1, D_2\}$. 
According to the relationship above, learning from the stitched sequence may help us obtain lower estimation error than learning from the separate two sequences. 

Note that for memoryless models like Poisson processes, such a stitching-based learning strategy will not cause model misspecification because the stitched sequence obeys the same model as that of the original sequences. 
However, for a more-complicated model like Hawkes processes or self-correcting processes, the stitching operation may introduce nonexistent triggering patterns. 
In such a situation, our stitching-based learning strategy suppresses the influence of warping function while raising the risk of model misspecification. 
Fortunately, as discussed in~\cite{xu17b}, when the intensity function is exponential-like function, the model misspecification problem is ignorable with a small number of stitching operations. 
The experiments in the experimental section further verifies the feasibility of this method.

\section{Related Work}
\subsection{Temporal point processes}
Point processes have proven to be useful in many applications, $e.g.$, financial analysis~\cite{bacry2012non} and social network analysis~\cite{zhou2013learning,zhao2015seismic}. 
However, most existing work does not consider learning parametric point processes from idiosyncratic observations with latent sample-specific effects. 
The methods in~\cite{lewis2011nonparametric,yan2015machine} try to estimate time scaling parameters for point process models, but they are only available for Hawkes processes whose event sequences share the same linear transformation of time, which cannot capture personalized and nonlinear phenomena. 
The work in~\cite{luo2015multi} is able to jointly learn different Hawkes processes by multi-task learning, but it does not register its learning results or learn sample-specific warping functions. 

\subsection{Data registration and model registration}\label{ssec:relate}
The idiosyncratic aspects of sequential data may be viewed in terms of a sample-specific ``warping'' of a common latent phenomena, which can be registered based on certain transformations. 
Typical methods include the dynamic time warping (DTW)~\cite{berndt1994using,moeckel1997measuring} and its variants~\cite{wang2016graphical,cuturi2017soft,ramsay1998curve}, the self-modeling registration method (SMR)~\cite{gervini2004self}, the moment-based method (MBM)~\cite{james2007curve}, the pairwise curve synchronization method (PACE)~\cite{tang2008pairwise}, and the functional convex averaging (FCA) method~\cite{liu2004functional}. 
These methods can be categorized in the same framework -- the registered curves and the corresponding warping functions are learned alternatively based on a nonlinear least-squares criterion. 
Instead of using the Euclidean metric, the work in~\cite{srivastava2011registration} obtains better data registration results by using the Fisher-Rao metric (FRM).
For those nonparametric models like Gaussian processes, warping data is beneficial to improve the robustness of learning methods~\cite{snelson2004warped,cunningham2012gaussian,snoek2014input,herlands2016scalable}.

The work in~\cite{panaretos2016amplitude,zemel2017fr} proposes a model-registration method. 
Specifically, the unregistered distributions of warped observations are first estimated by nonparametric models, and then the registered point process are estimated as the barycenter of the distributions in Wasserstein space~\cite{muskulus2011wasserstein}. 
Finally, the warping function between any unregistered distribution and the registered one is learned as an optimal transport~\cite{anderes2016discrete}. 
However, all of these methods focus on warping/unwarping continuous curves in a nonparametric manner, which are hard to register parametric point processes from idiosyncratic event sequences. 
The recent combination of Wasserstein learning and neural networks~\cite{arjovsky2017wasserstein,xiao2017wasserstein} achieves encouraging improvements on learning robust generative models from imperfect observations. 
However, the neural network-based model requires many time-consuming simulation steps in the learning phase, and cannot in general learn explicit warping functions.

\begin{figure*}[t]
\centering
\subfigure[$W(t)$]{
\includegraphics[height=0.165\linewidth]{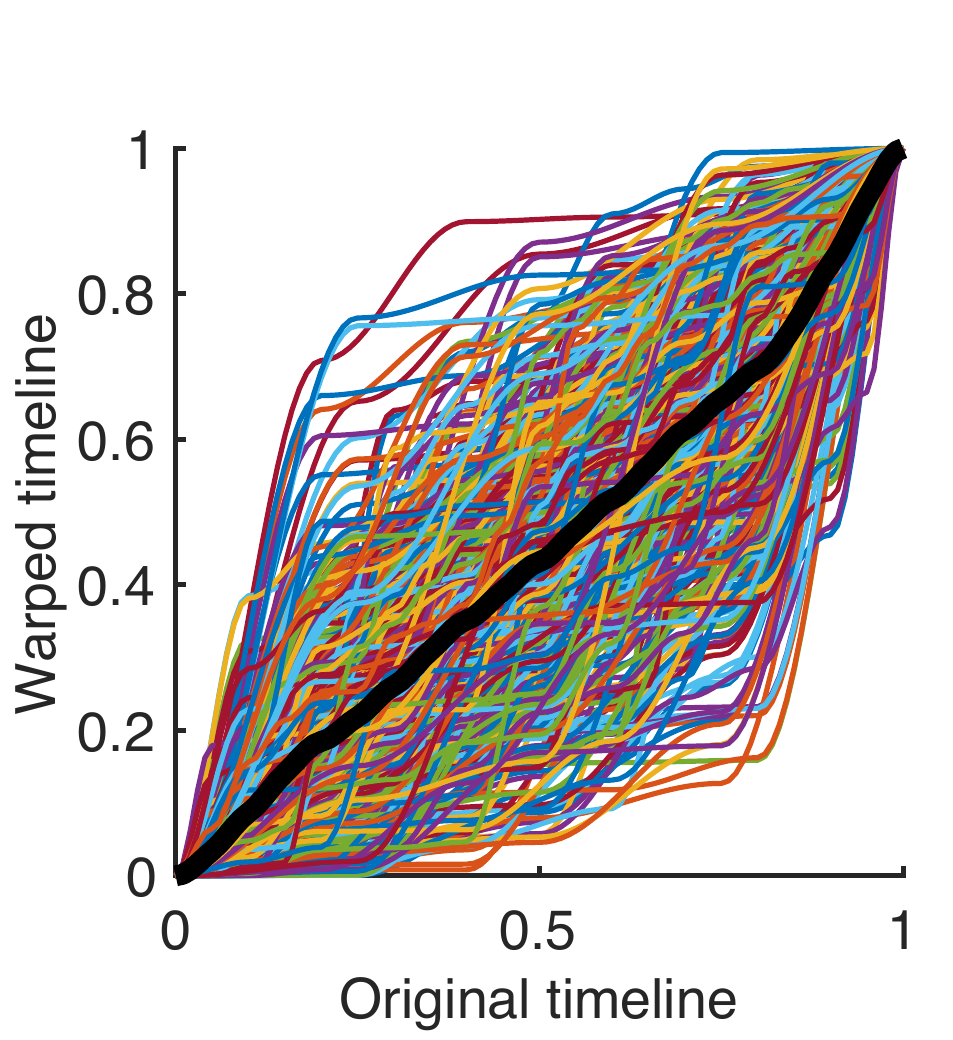}\label{warpfuncs}}
\subfigure[Inhomogeneous Poisson process]{
\includegraphics[width=0.33\linewidth]{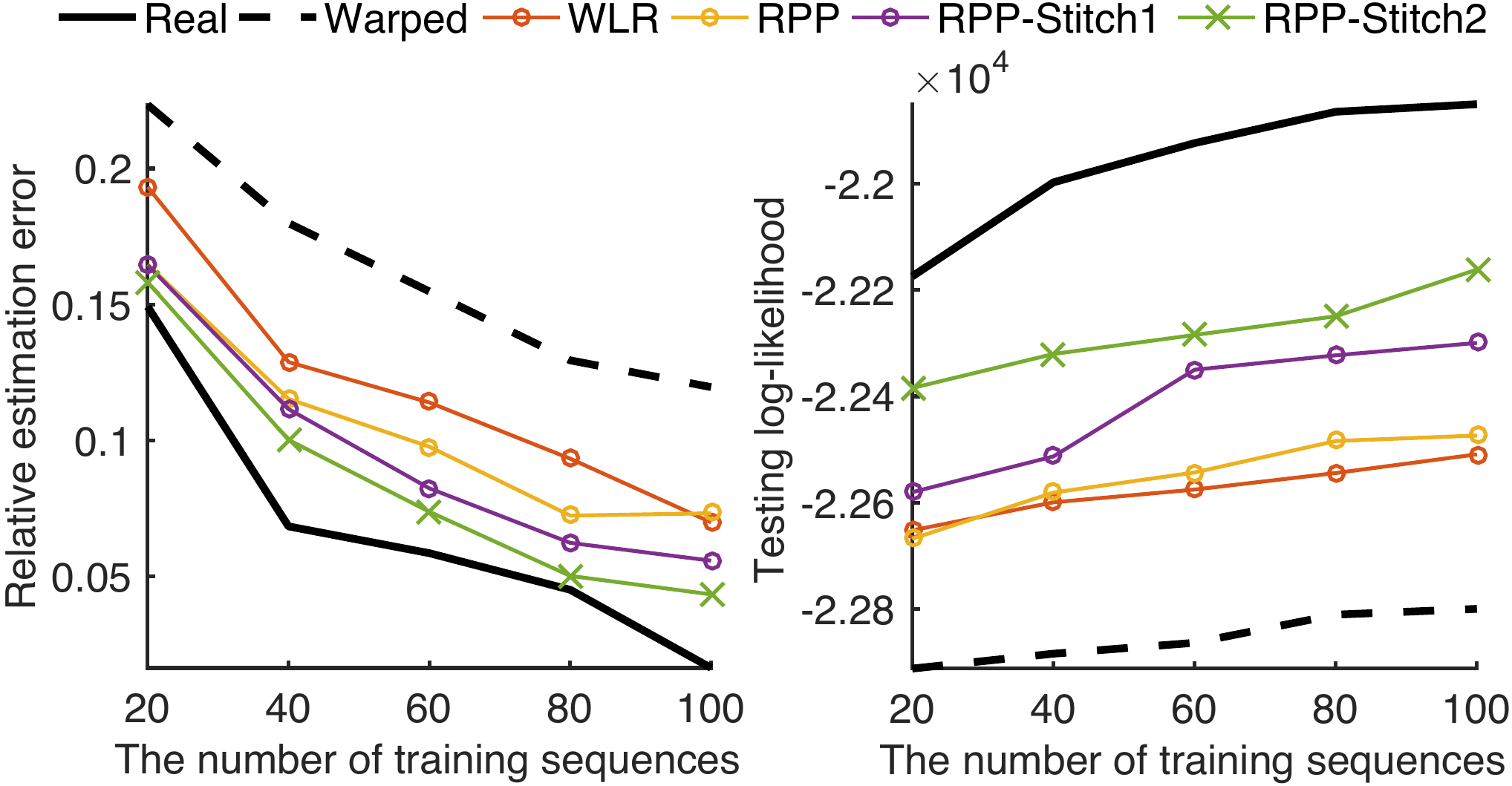}\label{synres1}}
\subfigure[Hawkes process]{
\includegraphics[width=0.33\linewidth]{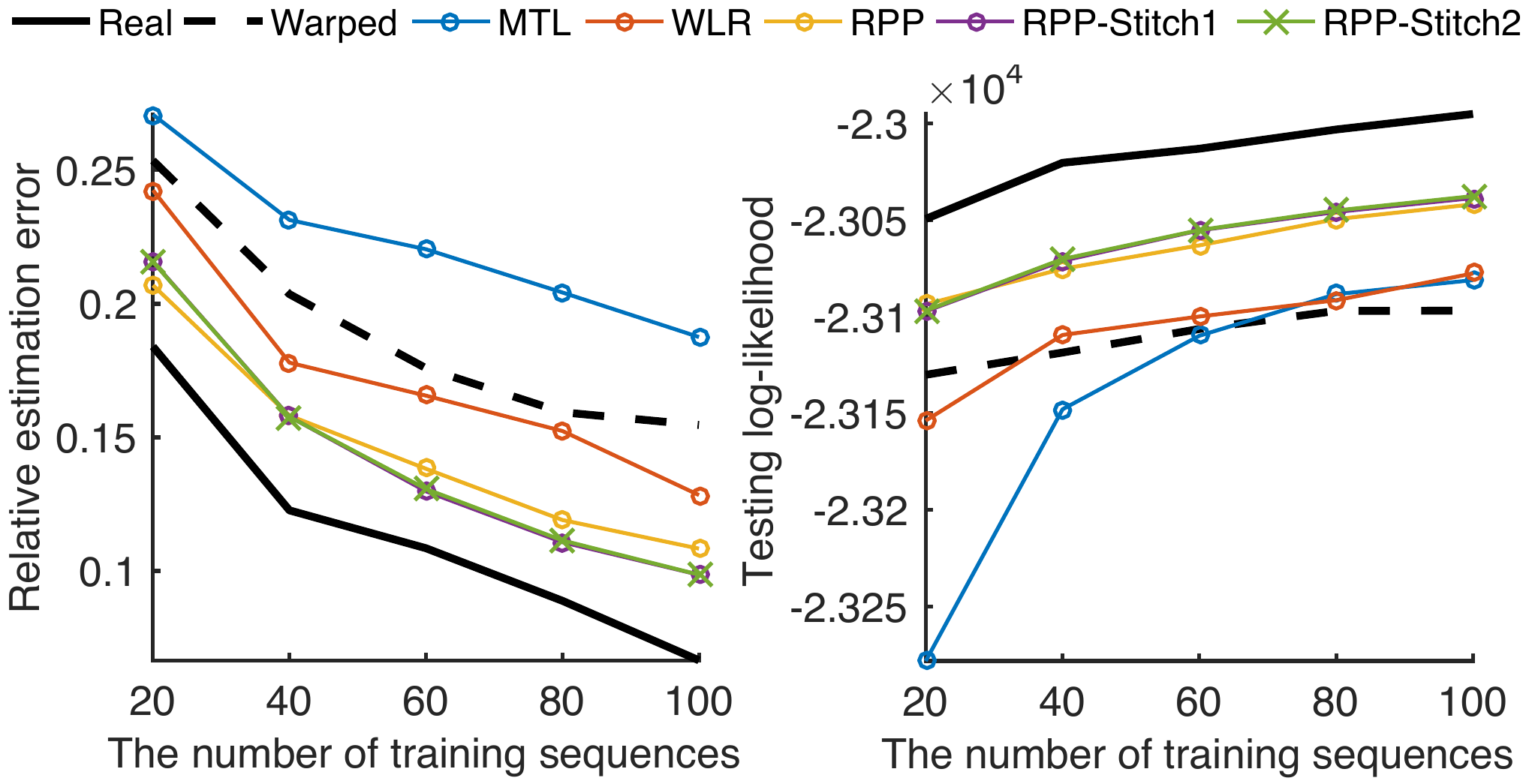}\label{synres2}}
\subfigure[Convergence]{
\includegraphics[height=0.165\linewidth]{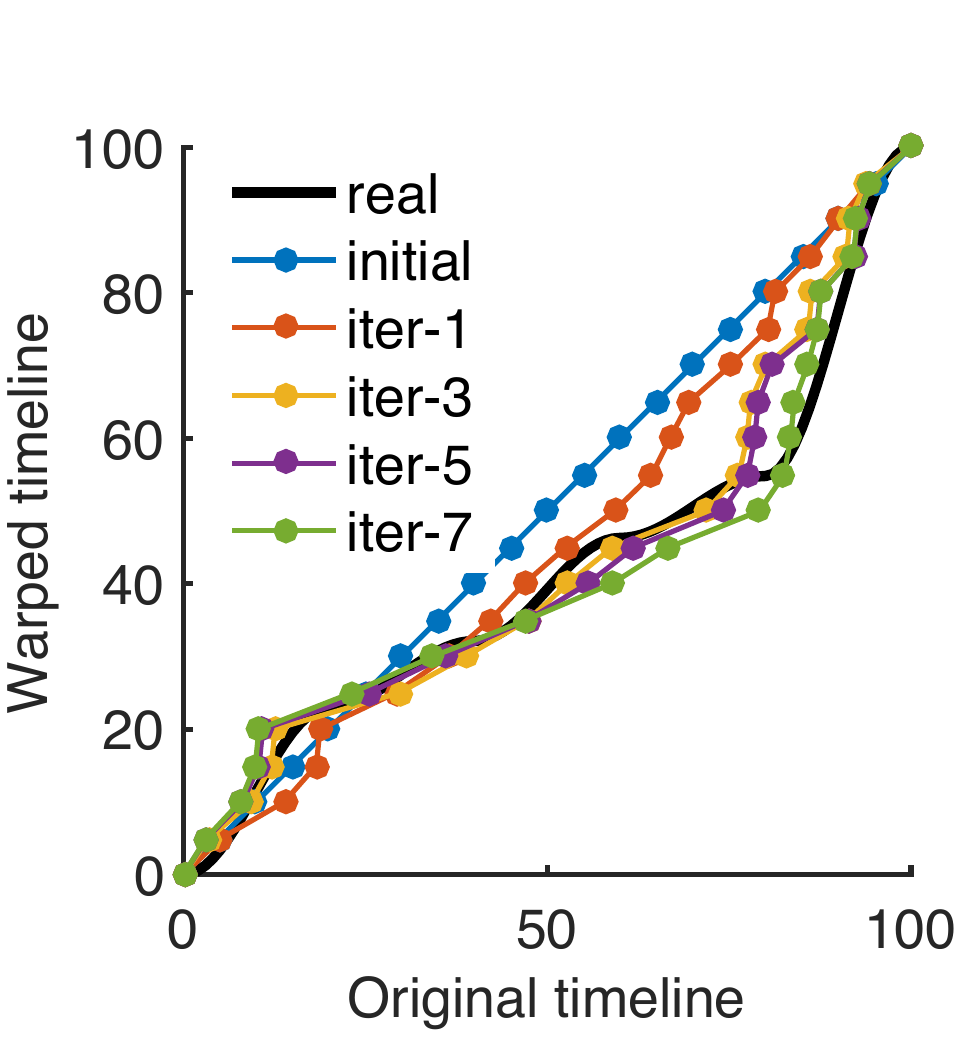}\label{conv}}
\vspace{-10pt}
\caption{\small Comparisons for various methods on synthetic data.\label{fig:synthetic}}
\end{figure*}

\section{Experiments}
Denote our point process registering method and its variant, assisted with the stitching operation, as \textbf{RPP} and \textbf{RPP-stitch}, respectively.
To demonstrate the feasibility and effectiveness of the proposed methods, we compare them to existing point process learning and registration methods, on both synthetic and real-world datasets. 
We compare to the following methods: purely maximum likelihood estimation based on warped observations (\textbf{Warped}), the multi-task learning-based method (\textbf{MTL})~\cite{luo2015multi}, and the Wasserstein learning-based registration method (\textbf{WLR})~\cite{panaretos2016amplitude}. 
Specifically, the MTL method learns specific parametric point processes jointly from warped event sequences with low-rank and sparse regularizers, and averages the learned parameters over all event sequences in Euclidean space. 
The WLR is the state-of-the-art model registration method focusing on point processes and their warped event sequences.  
To apply the WLR method to learn parametric point process models, we first follow the work in~\cite{panaretos2016amplitude}, learning the densities of observed events by kernel density estimation (KDE)~\cite{sheather1991reliable}, and learning the warping functions by finding the optimal transport between the densities and their barycenter in the Wasserstein space. 
Finally, we apply the reversed warping functions to unwarp the observations and learn a parametric point process. 

\subsection{Synthetic data}
We simulate a 1D inhomogeneous Poisson process and a $4$-dimensional Hawkes process. 
For each synthetic data set, we generate $200$ event sequences in the time window $[0,100]$ using Ogata's thinning method~\cite{ogata1981lewis} and divide them equally into a training set and a testing set. 
The intensity function of the Poisson process is represented as $\sum_{j=1}^5\exp_{t_j}(-(t-t_j))$, where $t_j$ is uniformly sampled from $[0,100]$, while the intensity function of the Hawkes process is defined as the model in~\cite{zhou2013learning}.
Each sequence in the training set is modified by a specific warping function. 
The warping functions are visualized in Fig.~\ref{warpfuncs}, in which each color curve represents a warping function and the black bold curve represents the average of all the functions. 
The generation method of the warping functions is given in Appendix~\ref{ap5}; 
it ensures that both the warping and the unwarping functions are monotone increasing and the averaged warping and unwarping functions are close to an identity function.

Given the training data, we can learn registered point process models by different methods and evaluate their performance on 1) the relative estimation error, and 2) the log-likelihood of testing set. 
For each method, we test with $5$ trials on the two data sets, and visualize its averaged results in Figs.~\ref{synres1} and~\ref{synres2}. 
The black bold curves correspond to the MLE based on unwarped data, which achieves the best performance ($i.e.$, the lowest estimation error and the highest log-likelihood), while the black dot curves correspond to the MLE based on warped data. 
The performance of a good registration method should be much better than the black dotted curves and approach to the block bold curves. 
Our RPP method achieves superior performance to MTL and WLR.\footnote{Because MTL is designed for Hawkes processes, we do not use it for Poisson processes.} 
The performance of MTL is even worse than that of applying MLE to warped data directly, especially in the case with few training data. 
This result implies that 1) the sparse and low-rank structure imposed in the multi-task learning phase cannot reflect the actual influence of warped data on the distribution of parameters, and 2) the average of the parameters in the Euclidean space does not converge well to the ground truth. 
The performance of WLR is comparable to that of applying MLE to warped data directly, which verifies our claim that the WLR is unsuitable for learning complicated point processes when observations are not sufficient.

Both MTL and WLR rely on a strategy of learning a specific model for each event sequence, and then averaging the models in a predefined space. 
This strategy ignores a fact that the number of events in a single event sequence is often insufficient to learn a reliable model in practice. 
Our RPP method, by contrast, learns a single registered model and all warping functions jointly in an iterative manner, rather than in independent steps. 
As a result, our method suppresses the risk of over-fitting and achieves improved results. 
Further, we illustrate the learning process of a warping function in Fig.~\ref{conv} and verify the convergence of our RPP method. 
The black bold curve corresponds to the ground truth and the blue line is the initialization of our estimation. 
Applying our RPP method, the learning result converges after $7$ iterations, and the final estimation of the warping function approaches the ground truth.

We also examine the usefulness of the stitching strategy. 
In particular, in Fig. \ref{fig:synthetic} ``RPP-Stitch $K$'' denotes that for each event sequence, we randomly stitch it with $K$ other event sequences, and then apply our RPP method to the $200$ stitched sequences in time window $[0, 100(K+1)]$. 
We can find that for both Poisson processes and Hawkes processes, ``RPP-Stitch 1'' obtains better results than original RPP method, which verifies the improvements caused by the stitching strategy. 
However, for Poisson processes the improvements can be further enhanced by applying stitching operations multiple times ($i.e.$, $K=2$), while for Hawkes processes the improvements are almost unchanged. 
As we discussed in Section~\ref{sec:stitch}, applying too many stitching operations to the point processes with history-dependent intensity functions may cause model misspecification and counteract the benefits from suppressing distortions.

\subsection{Real-world data}
We test our methods, and compare with the WLR, on two real-world datasets: the MIMIC III dataset~\cite{johnson2016mimic}, and the Linkedin dataset~\cite{xu17b}. 
The MIMIC III dataset contains over ten thousand patient admission records over ten years. 
Each admission record is a sequence, with admission time stamps and the ICD-9 codes of diseases. 
Following~\cite{xu17b}, we assume that there are triggering patterns between different diseases, which can be modeled by a Hawkes process. 
We focus on modeling the triggering patterns between the diseases of the circulatory system, which are grouped into $8$ categories. 
We extract $1,129$ admission records related to the $8$ categories as the training set. 
Each record can be viewed as an event sequence warped from a ``standard'' record, because of the idiosyncratic nature of different patients.
For the Linkedin dataset, we extract $709$ users having working experience in $7$ IT companies. 
Similarly, the timeline of different users can be different, because they have different working experience and personal conditions, and the status of the job market when they jump is different as well. 
We want to learn a ``standard'' Hawkes process to measure the relationships among the companies and exclude these uncertain factors.

We apply different model registration methods to learn registered Hawkes processes from the two real-world datasets.
The evaluation is challenging because both the groundtruth of the model and that of the warping functions are unknown. 
Fortunately, we can use learning results to evaluate the risks of under- and over-registration for different methods in an empirical manner. 
Given unwarped event sequences estimated by different methods, we learn the parameter of model $\theta^{*}$ and estimate its variance $var(\theta^*)$ by parametric bootstrapping~\cite{wassermann2006all}. 
For the method with a lower risk of under-registration, its learning result should be more stable and the estimated variance should be smaller. 
Therefore, we can use the estimated variance as a metric for the risk of under-registration, $i.e.$, $risk_{under}=var(\theta^*)$.
We define the following metric to evaluate the risk of over-registration:
$risk_{over}=\frac{\int_{0}^{T}|s - \overline{W}(s)|^2ds}{\frac{1}{M}\sum_{m=1}^{M}\int_{0}^{T}|W_m(s) - \overline{W}(s)|^2ds}$, 
where $\overline{W}(s)=\frac{1}{M}\sum_{m}W_{m}(s)$.
The numerator is the distance between the mean of warping functions and an identity function, and the denominator is the variance of warping functions. 
When the estimated warping functions have a small variance ($i.e.$, the warping functions are similar to each other) but are very distinct from identity function ($i.e.$, the bias of the warping functions is large), it means that the corresponding method causes over-registration. 

\begin{figure*}[t]
\centering
\begin{minipage}[b]{0.18\linewidth}
\subfigure[MIMIC]{
\includegraphics[width=0.75\columnwidth]{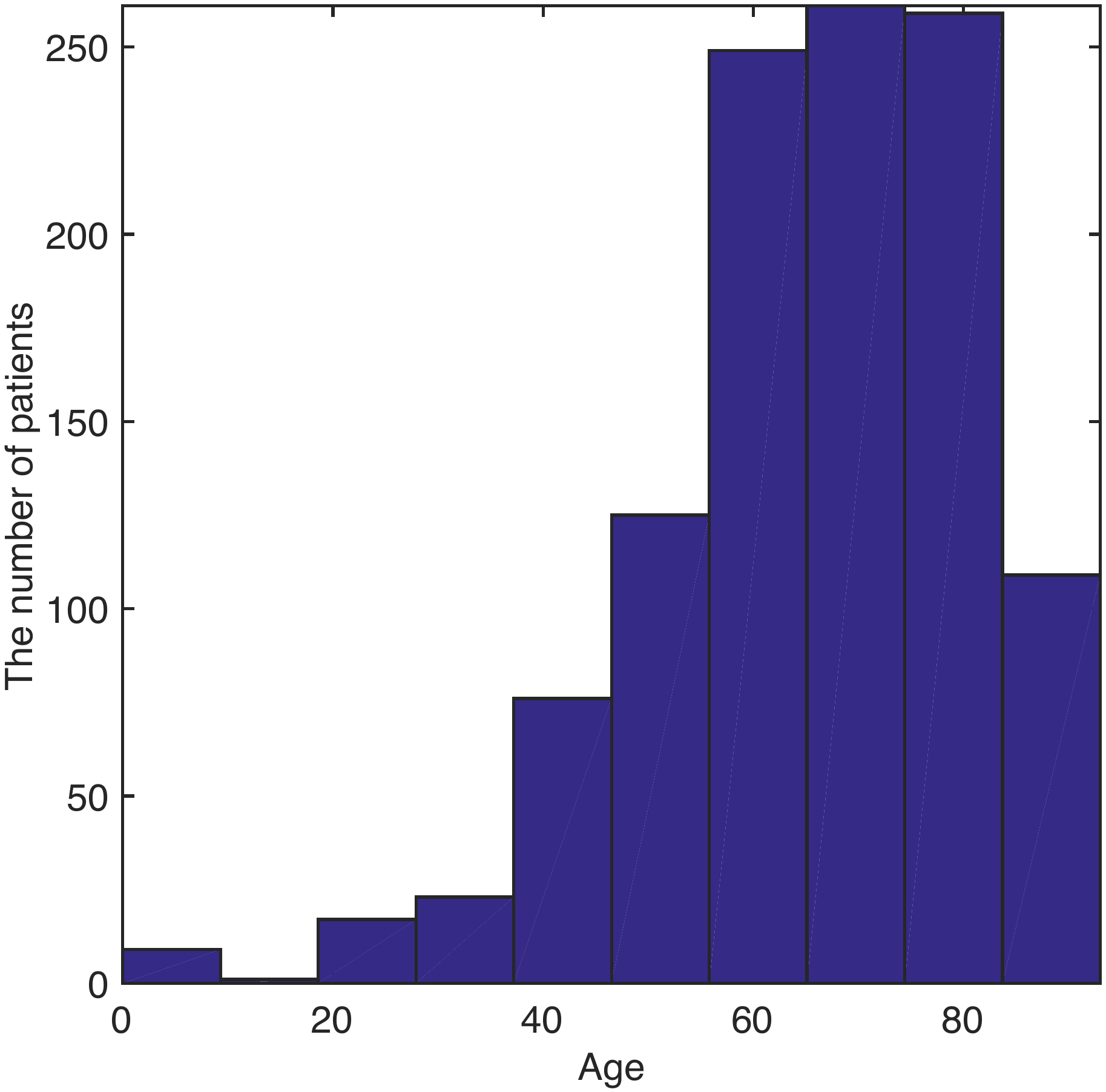}\label{hist1}}\\ \vspace{-0.5cm}
\subfigure[Linkedin]{
\includegraphics[width=0.75\columnwidth]{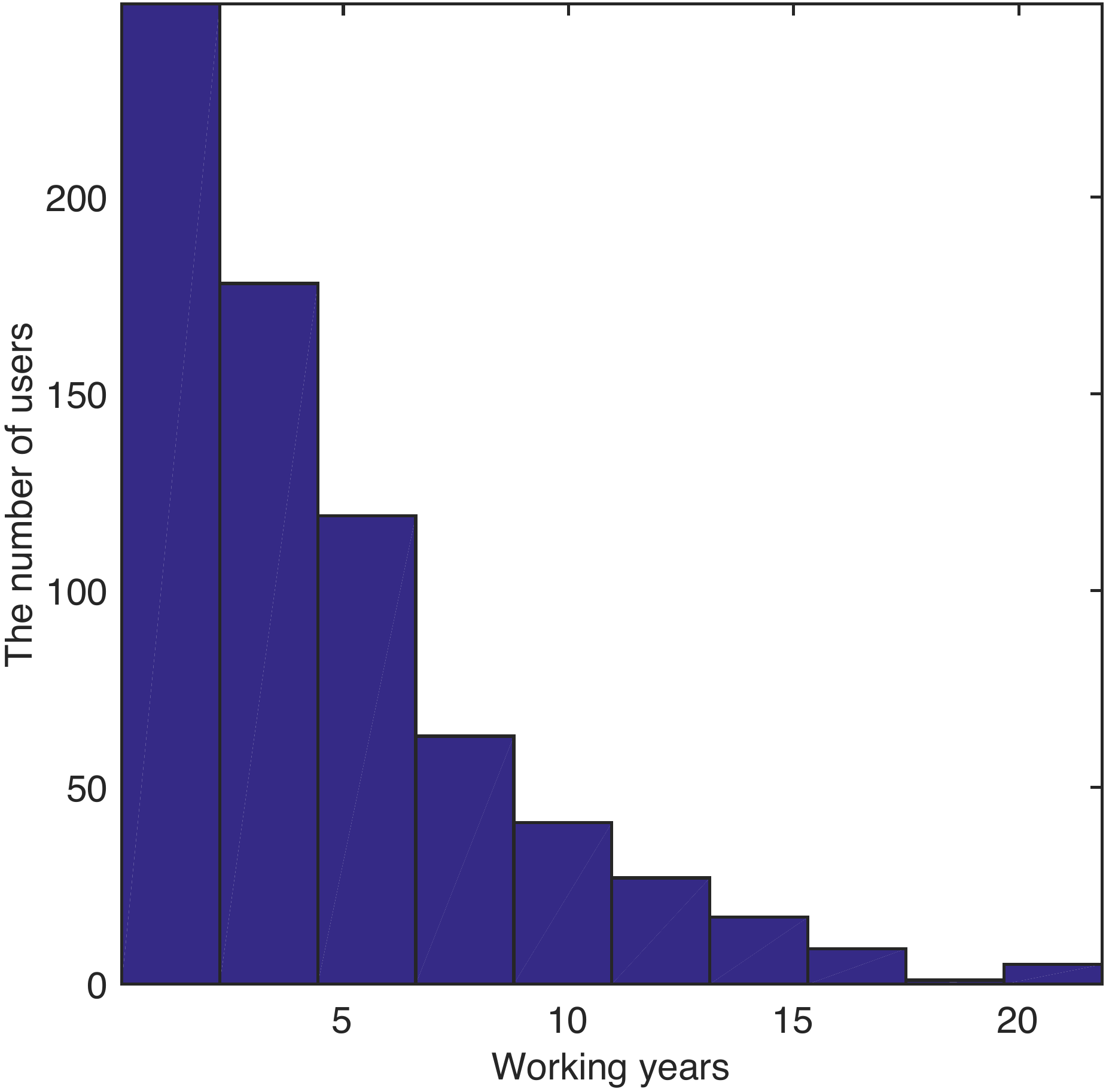}\label{hist2}}
\end{minipage}
\subfigure[MIMIC III]{
\includegraphics[width=0.39\linewidth]{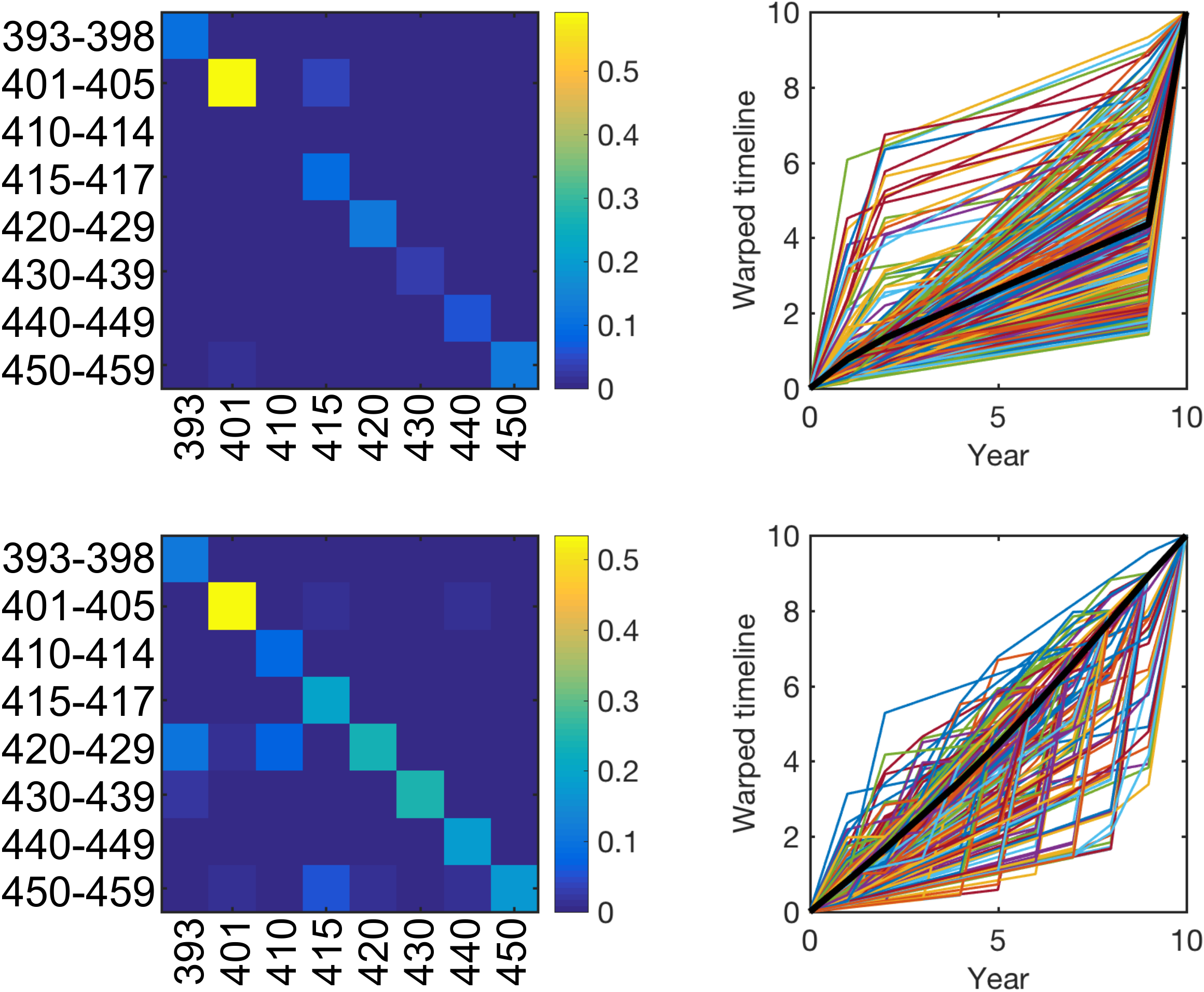}\label{result1}}
\subfigure[Linkedin]{
\includegraphics[width=0.39\linewidth]{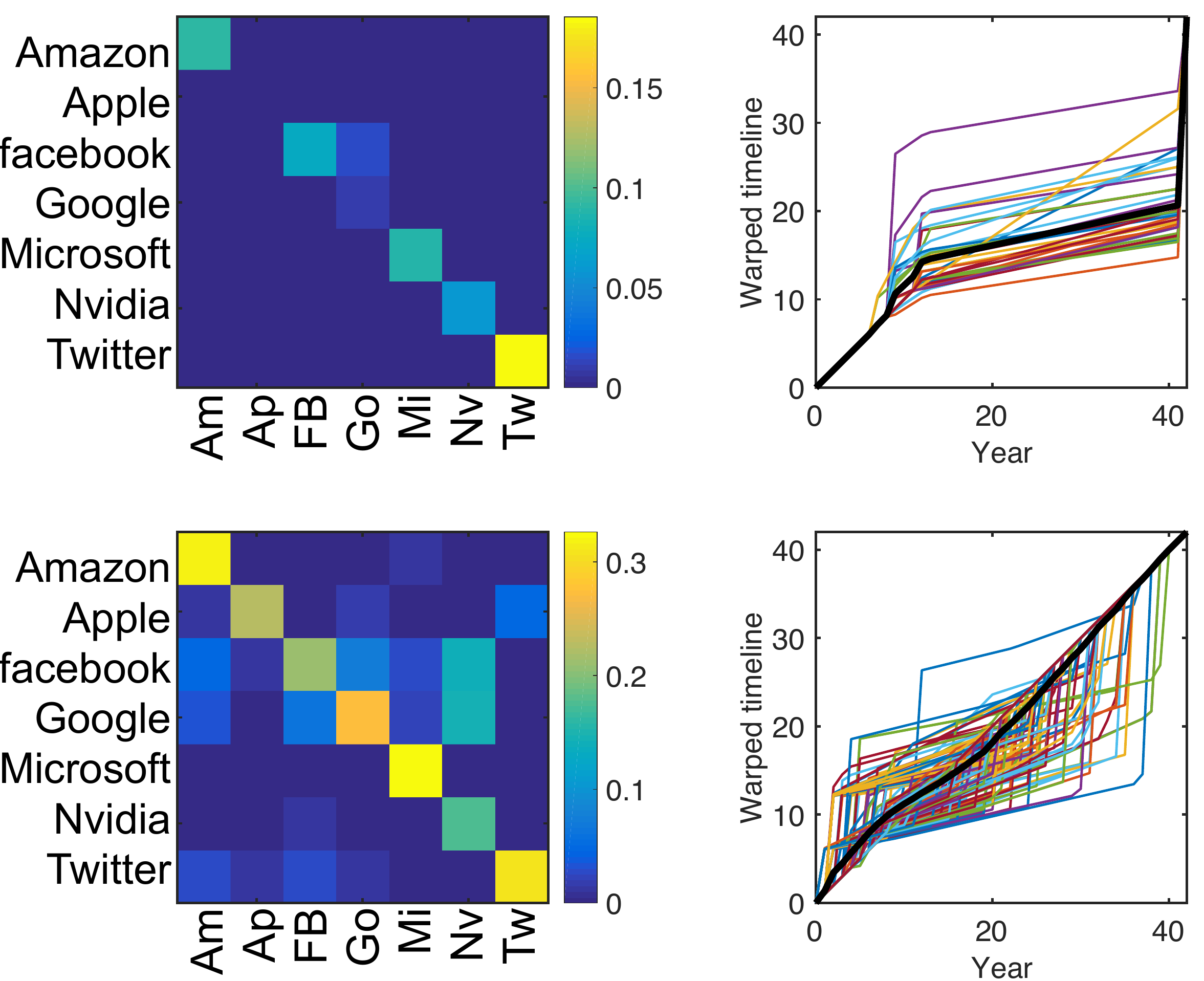}\label{result2}}\vspace{-10pt}
\caption{\small Experimental results of our method on real-world datasets. In (c) and (d), the first row corresponds to the infectivity matrix and the warping functions learned by WLR, and the second row corresponds to those learned by our RPP-Stitch1. The black bold curves are the average of warping functions.}
\end{figure*}

The side information of the dataset is also helpful to evaluate the appropriateness of the learning result. 
In Fig.~\ref{hist1}, most of the admission records in the MIMIC III dataset are from relatively old patients.
The incidence of circulatory system diseases is mainly correlated with patient age. 
Learning a ``standard'' patient model from a dataset dominated by old patients, we can imagine that the admission record of an old patient should be more similar to that of the ``standard'' patient, and the corresponding warping function should be closer to the identity function. 
Therefore, given the deviations between learned warping functions and the identity function, we can calculate the Kendall's rank correlation between the warping deviations and the ages of the patients. 
Similarly, in Fig.~\ref{hist2}, most of samples in the Linkedin dataset are from young users with $4$ or fewer working years, so these young users' behaviors should likely be close to that of the ``standard'' job-hopping model learned from the data, and the warping deviations should be correlated with the working years. 

\begin{table}[t]
  \centering
  \caption{\small Comparisons for various methods on two real-world data.\label{tab1}}%\vspace{-5pt}
    \begin{small}
      \begin{tabular}{
        @{\hspace{3pt}}c@{\hspace{3pt}}|
        @{\hspace{3pt}}c@{\hspace{3pt}}|
        @{\hspace{3pt}}c@{\hspace{3pt}}
        @{\hspace{3pt}}c@{\hspace{3pt}}
        @{\hspace{3pt}}c@{\hspace{3pt}}
        }
        \hline\hline
        Data &Method & $risk_{under}$ & $risk_{over}$ & Rank Corr.\\ 
        \hline
        \multirow{3}{*}{MIMIC-III}& WLR & 0.018 & 0.055 & 0.025\\
        & RPP & 0.011 & 0.009 & \textbf{0.053}\\ 
        & RPP-Stitch1 & \textbf{0.003} &\textbf{0.002} &\textbf{0.053}\\
        \hline
        \multirow{3}{*}{LinkedIn}& WLR & 0.029 & 0.657 & 0.344\\
        & RPP & 0.025 & 0.010 & 0.375\\
        & RPP-Stitch1 & \textbf{0.005} &\textbf{0.006} &\textbf{0.387}\\
        \hline\hline
      \end{tabular}
    \end{small}
\end{table}

Table~\ref{tab1} shows the comparison between our methods (RPP and RPP-Stitch1) and the WLR method on these two datasets. 
We find that our RPP method outperforms WLR consistently on different metrics and different datasets, obtaining lower risks of under- and over-registration and higher rank correlation. 
In particular, the low risk of under-registration means that the parameter $\theta^*$ learned by our method is stable. 
The low risk of over-registration means that the warping/unwarping functions we learned have good diversity and low bias. 
The high rank correlation verifies the justifiability of our method -- the warping deviations of dominant samples ($i.e.$, the old patients in MIMIC III and young employees in Linkedin data) are smaller than those of minor samples ($i.e.$, the young patients and the old employees). 
Similar to the case of synthetic data, applying the stitching strategy once, we further improve the learning results.

Figures~\ref{result1} and~\ref{result2} compare the infectivity matrices\footnote{The infectivity matrix is denoted $\bm{\Psi}=[\psi_{cc'}]$. Its element is the integral of impact function over time, $i.e.$, $\psi_{cc'}=\int_{0}^{T}\phi_{cc'}(s)ds$.} of the registered Hawkes processes and the warping functions learned by WLR and our RPP-Stitch1 for the two datasets. 
These results further verify the effectiveness of the proposed method. 
First, the warping/unwarping functions we learned have good diversity and the bias of the functions is lower than than that of the functions learned by WLR. 
Second, the infectivity matrices learned by our RPP-Stitch1 are more dense and informative, which reflect some reasonable phenomena that are not found by WLR. 
For the MIMIC III data, the infectivity matrix of WLR only reflects the self-triggering patterns of the disease categories, while ours is more informative: the $5$-th row of our matrix (the bottom-left subfigure in Fig.~\ref{result1}) corresponds to the category ``other forms of heart disease'' (ICD-9 code 420-429), which contains many miscellaneous heart diseases and has complicated relationships with other categories. 
Our learning result reflects this fact -- the $5$-th row of our infectivity matrix contains many non-zero elements.
For the Linkedin data, the infectivity matrix of our method reveals more information besides the self-triggering patterns: 
1) The values of ``Facebook-Google'' and ``Google-Facebook'' imply that job-hopping behaviors happen frequently between Facebook and Google, which reflects fierce competition between these companies. 
2) The values of ``Facebook-Nvidia'' and ``Google-Nvidia'' reflect the fact that recent years many Nvidia's employees moved to Google and Facebook to develop the hardware of AI.  
More detailed analyses are given in Appendix~\ref{ap6}.

\begin{figure}[t]
\centering
\subfigure[$\gamma$ from $10^{-3}$ to $10$]{
\includegraphics[width=0.4\linewidth]{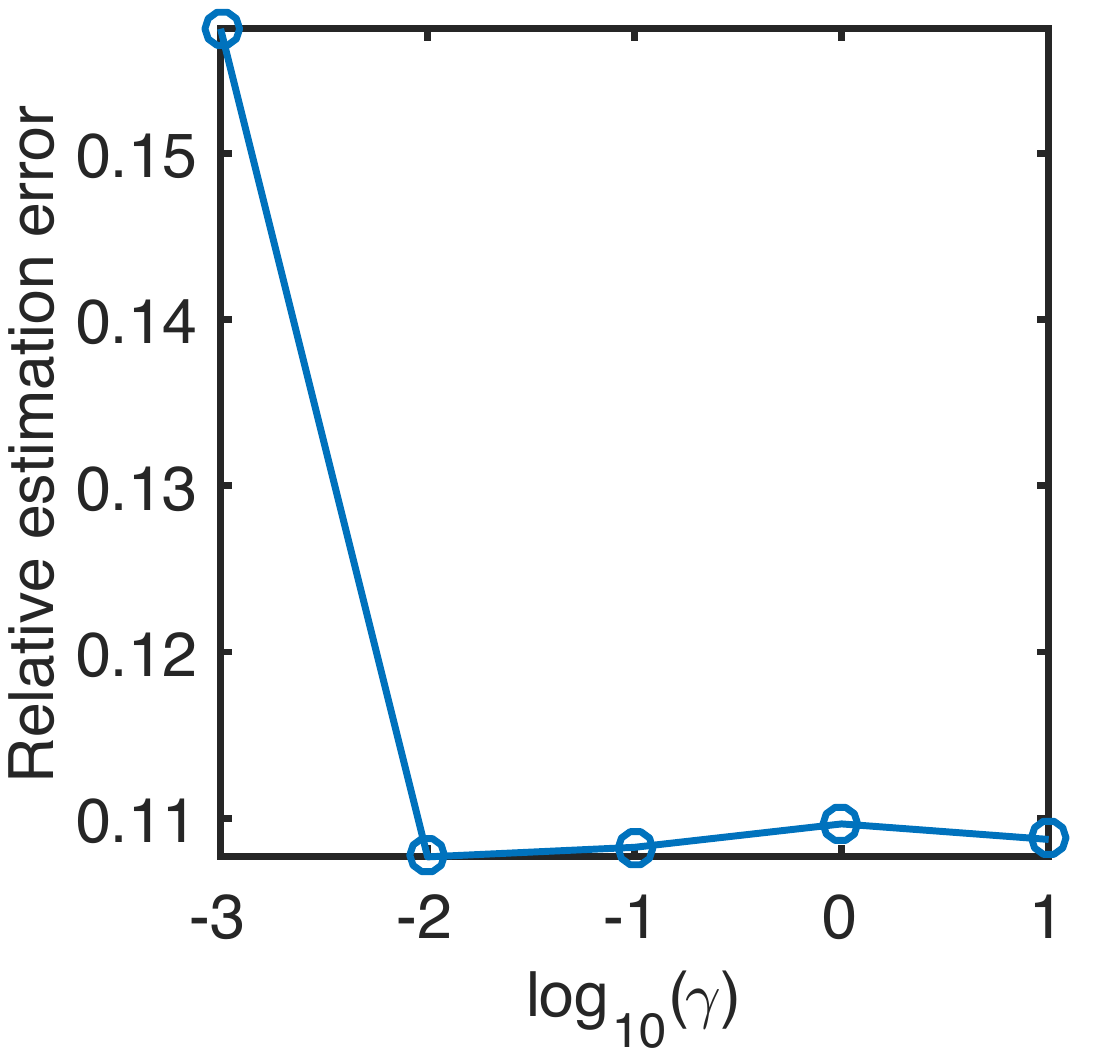}\label{gamma}}~
\subfigure[$L$ from $5$ to $100$]{
\includegraphics[width=0.4\linewidth]{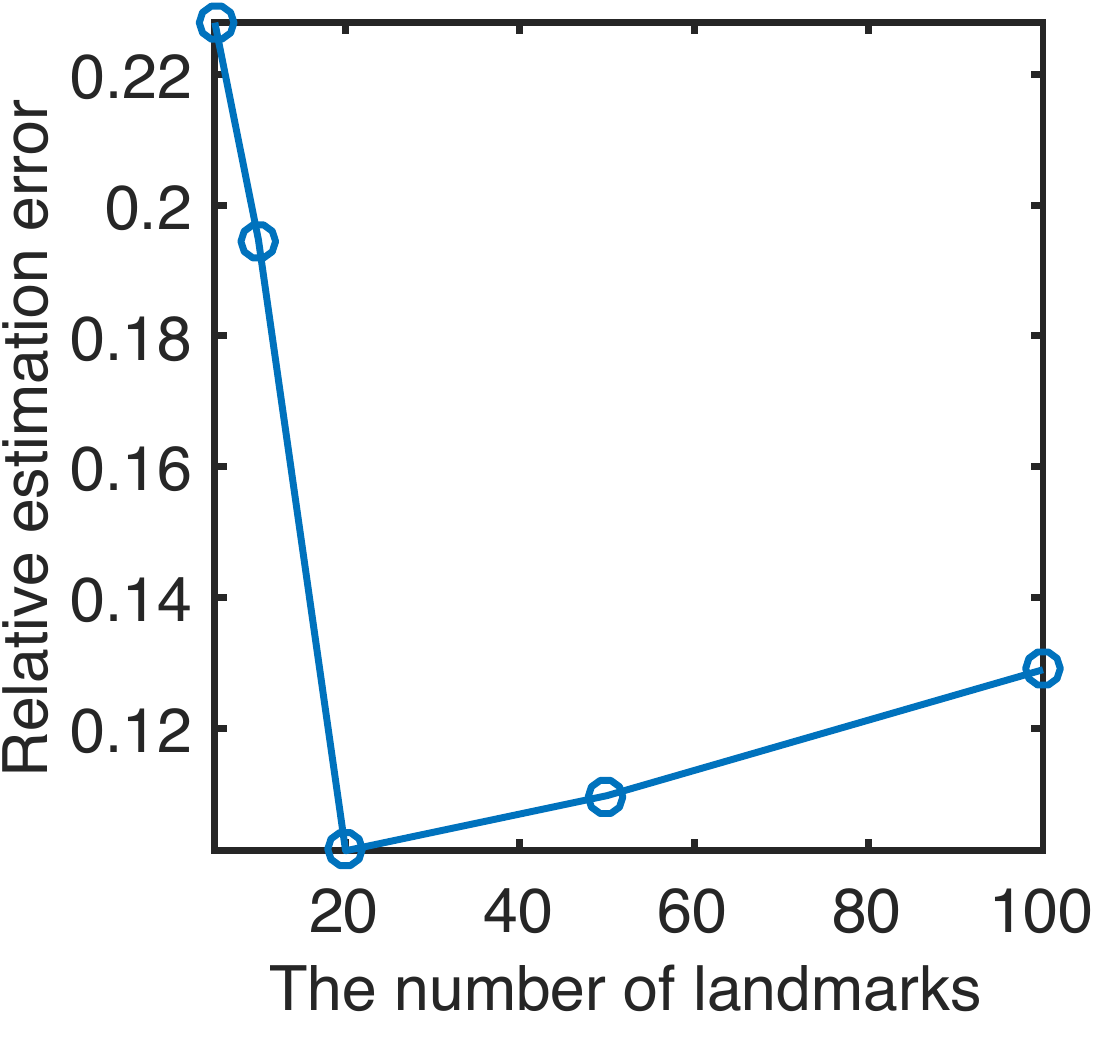}\label{landmark}}\vspace{-10pt}
\caption{\small Illustration of robustness. The relative estimation errors with respect to the changes of $\gamma$ and $L$ are shown respectively.
We can find that for both $\gamma$ and $L$, the relative estimation errors keep stable in the wide range.\label{robust}}
\end{figure}

\subsection{Robustness analysis}
We investigate the robustness of our method to variations in its parameters, including the weight of regularizer $\gamma$ and the number of landmarks $L$. 
In particular, we learn models from the synthetic data by the proposed method with different configurations, and visualize the estimation errors with respect to these two parameters in Fig.~\ref{robust}. 
The weight $\gamma$ controls the importance of the regularizer, which is correlated with the strictness of the unbiasedness assumption.
The larger $\gamma$, the more similarity we have between unwarping function and identity function.
In Fig.~\ref{gamma} we find that our method is robust to the change of $\gamma$ in a wide range ($i.e.$, from $10^{-3}$ to $1$). 
When $\gamma$ is too small ($i.e.$, $\gamma=10^{-3}$), however, the estimation error increases because the regularizer is too weak to prevent over-registration. 
The number of landmarks $L$ has an effect on the representation power of our method. 
In Fig.~\ref{landmark}, we find that the lowest estimation error is achieved when the number of landmarks $L=20$. 
When $L$ is too small, our piecewise linear model is over-simplified and cannot fit complicated warping functions well.
When $L$ is too large, (\ref{opt:ab}) has too many variables and the updating of warping function suffers to the problem of over-fitting.

\section{Conclusions and Future work}
We have proposed an alternating optimization method to learn parametric point processes from idiosyncratic observations. 
We demonstrate its justifiably and advantages relative to existing methods. 
Additionally, we also consider the influence of the stitching operation on the learning results and show the potential benefits empirically. 
Our method has potential for many applications, including admission data analysis and job-hopping behavior analysis.
In the future, we plan to extend our method to more complicated point process models and analyze the influence of the stitching operation theoretically.

%\section{Acknowledgment}
%We thank Yoav Zemel for discussions and comments about this work.

\bibliography{example_paper}
\bibliographystyle{icml2018}
%\clearpage
\newpage

\section{Appendix}
\subsection{Exponential-like intensity functions}\label{ap1}
We given some typical and important point processes with exponential-like intensity functions, $i.e.$, $\lambda(t)=\sum_j\exp_{t_j}(f(t;\theta,\mathcal{H}_t^{\mathcal{C}}))$. 
More specifically, for Hawkes processes and self-correcting processes, this formulation can be further rewritten as $\lambda(t)=\sum_j \alpha_j\exp(\beta_j t)$. 
For the convenience of expression, we only consider 1-D point processes, $i.e.$, the number of event types $C=1$, but these examples can be easily extended to multi-dimensional cases. 

\textbf{Hawkes processes.} The intensity function of a 1-D Hawkes process is
\begin{eqnarray}\label{hp2}
\begin{aligned}
\lambda(t)=\mu + \sideset{}{_{t_i<t}}\sum \phi(t-t_i),
\end{aligned}
\end{eqnarray}
A typical implementation of the impact function $\phi(t)$ is exponential function, $i.e.$, $\rho\exp(-wt)$ in~\cite{hawkes1974cluster,lewis2011nonparametric,zhou2013learning,yan2015machine}. 
Therefore, we can rewrite (\ref{hp2}) as
\begin{eqnarray}\label{hp:exp}
\begin{aligned}
\lambda(t)&=\mu + \sideset{}{_{t_i<t}}\sum \phi(t-t_i)\\
&=\mu\exp(0t) + \sideset{}{_{t_i<t}}\sum \rho\exp(wt_i)\exp(-wt))\\
&=\sideset{}{_{j=1}^{J}}\sum\alpha_j\exp(-\beta_j t),
\end{aligned}
\end{eqnarray}
where $J=1+|\{t_i:~t_i<t\}|$. 
We can find that for $j=1$, $\beta_j=0$ and $\alpha_j=\mu$; for $j=2,..,J$, $\beta_j=w$ and $\alpha_j=\rho\exp(wt_i)$. 

\textbf{Self-correcting processes.} The intensity function of a 1-D self-correcting process~\cite{isham1979self,xu2015trailer} is
\begin{eqnarray}\label{sc2}
\begin{aligned}
\lambda(t)=\exp(\mu t - \sideset{}{_{t_i<t}}\sum \phi(t_i)).
\end{aligned}
\end{eqnarray}
Generally, $\phi(t)$ can be 1) a linear function of time, $i.e.$, $\phi(t)=\rho t$; or 2) a constant, $i.e.$, $\phi(t)=\rho$. 
In this case, we can simply represent $\lambda(t)$ as an exponential function $\alpha\exp(-\beta t)$, where $\alpha=\exp(-\sum_{t_i<t} \phi(t_i))$ and $\beta=-\mu$. 

\subsection{The proof of Theorem~\ref{thm1}}\label{ap7}

\begin{proof}
Denote an original (unwarped) event sequence as $D$. 
The negative log-likelihood function of the target point process $N_{\theta}$ can be written as
\begin{eqnarray}
\begin{aligned}
-\log\mathcal{L}(\theta;D)=\int_{0}^{T}\lambda(s)ds-\sum_{i}\log\lambda(t_i),
\end{aligned}
\end{eqnarray}
where $t_i$ is the $i$-th event of the sequence $D$. 
When the training sequence $D$ is warped by a warping function $W:~[0,T]\mapsto [0,T]$ and the warping function is continuous and  differentiable (almost everywhere), we have
\begin{eqnarray}
\begin{aligned}
&-\log\mathcal{L}(\theta;S)\\
=&\int_{0}^{T}\lambda(W(s))ds-\sum_{i}\log\lambda(W(t_i))\\
=&\int_{0}^{T}\lambda(s)dW^{-1}(s)-\sum_{i}\log\lambda(W(t_i)),
\end{aligned}
\end{eqnarray}
where $S$ is the warped data. 

\textbf{Sufficiency.} When the target point process is a homogeneous Poisson process, i.e., $\lambda(t)=\mu$, we can find that
\begin{eqnarray}
\begin{aligned}
-\log\mathcal{L}(\theta;S)
=-\log\mathcal{L}(\theta;D)
=T\mu - I\log\mu,
\end{aligned}
\end{eqnarray}
where $I$ is the number of events. 
Therefore, both $\hat{\theta}^*$ and $\hat{\theta}$ are equal to $\frac{I}{T}$. 

When we relax the range of $W(t)$ but assume that it is a translation, i.e., $W(t)=t+\tau$, the relative distance between arbitrary two events, i.e., $t_i-t_j = W(t_i)-W(t_j)$, is unchanged. 
Based on the stationarity of the target point process, the learning result is unchanged as well.

\textbf{Necessity.} 
When the target point process has exponential-like intensity function, the negative log-likelihood is a convex function of $\theta$. 
The warping function does not change the convexity of the negative log-likelihood. 
Therefore, when $\hat{\theta}^*=\hat{\theta}$, we have
\begin{eqnarray}\label{eqn}
\begin{aligned}
&\frac{\partial -\log\mathcal{L}(\theta;S)}{\partial\theta}\Bigr|_{\hat{\theta}^*}=0,
\end{aligned}
\end{eqnarray}
for the target point process. 

Even in the simplest case, i.e., the intensity is a single exponential function $\lambda(t)=\alpha_\theta\exp(-\beta t)$ and only $\alpha_\theta$ is a single coefficient related to the parameter $\theta$, we have
\begin{eqnarray*}
\begin{aligned}
&-\log\mathcal{L}(\theta;S)\\
=&-\log\mathcal{L}(\theta;D)+\int_{0}^{T}(1-(W^{-1})'(s))\lambda(s)ds\\
&-\sum_i \log\frac{\lambda(W(t_i))}{\lambda(t_i)}\\
=&-\log\mathcal{L}(\theta;D)+\alpha_\theta\int_{0}^{T}(1-(W^{-1})'(s))\exp(-\beta s)ds\\
&-\sum_i \log\frac{\exp(-\beta W(t_i))}{\exp(-\beta t_i)}.
\end{aligned}
\end{eqnarray*}
Here, we have
\begin{eqnarray}
\begin{aligned}
\frac{\partial -\log\mathcal{L}(\theta;D)}{\partial\theta}\Bigr|_{\hat{\theta}^*}=0,
\end{aligned}
\end{eqnarray}
and the last term $-\sum_i \log\frac{\exp(-\beta W(t_i))}{\exp(-\beta t_i)}$ is a constant with respect to $\theta$, 
therefore, $\frac{\partial -\log\mathcal{L}(\theta;S)}{\partial\theta}|_{\hat{\theta}^*}=0$ is equivalent to $\int_{0}^{T}(1-(W^{-1})'(s))\exp(-\beta s)ds\equiv 0$ for all kinds of event sequences. 
This condition satisfies in two situations: 1) $(W^{-1})'(s)\equiv 1$, which corresponds to a translation function; 2) $\beta=0$, such that $\lambda(t)=\alpha_\theta$ is a constant, which corresponds to a homogeneous Poisson process. 
\end{proof}

\subsection{The derivation of (\ref{surrogate})}\label{ap2}
Based on the assumption 3 of the target point process, the negative log-likelihood in (\ref{opt:wm}) can be rewrite as
\begin{eqnarray}\label{nll}
\begin{aligned}
&-\log \mathcal{L}(\theta^k; W_m^{-1}(S_m))\\
=&\sum_{c=1}^{C}\int_{0}^{T}\lambda_c(W_m^{-1}(s))ds-\sum_{i=1}^{I_m}\log\lambda_{c_i^m}(W_m^{-1}(t_i^m))\\
=&\sideset{}{_{c=1}^{C}}\sum\int_{W_m^{-1}(0)}^{W_m^{-1}(T)}\lambda_c(s)dW_m(s)\\
&-\sideset{}{_{i=1}^{I_m}}\sum\log\Biggl(\sideset{}{_{j=1}^{J_i}}\sum\alpha_j\exp(-\beta_j W_m^{-1}(t_i^m))\Biggr)\\
=&\sideset{}{_{c=1}^{C}}\sum\int_{0}^{T}\lambda_c(s)dW_m(s)\\
&-\sideset{}{_{i=1}^{I_m}}\sum\log\Biggl(\sideset{}{_{j=1}^{J_i}}\sum\alpha_j\exp(-\beta_j W_m^{-1}(t_i^m))\Biggr)\\
=&\mathcal{A}+\mathcal{B}.
\end{aligned}
\end{eqnarray}
On one hand, based on the piecewise linear model of $W_m^{-1}$, the term $\mathcal{A}$ can be further rewritten as
\begin{eqnarray}\label{termA}
\begin{aligned}
\mathcal{A}=&\sideset{}{_{c=1}^{C}}\sum\int_{W_m^{-1}(0)}^{W_m^{-1}(T)}\lambda_c(s)dW_m(s)\\
=&\sideset{}{_{c=1}^{C}}\sum\sideset{}{_{l=1}^{L-1}}\sum\int_{W_m^{-1}(t_l)}^{W_m^{-1}(t_{l+1})}\lambda_c(s)\frac{dW_m(s)}{ds}ds\\
=&\sideset{}{_{l=1}^{L-1}}\sum\underbrace{\frac{1}{a_l^m}}_{W_m^{\prime}}\underbrace{\sideset{}{_{c=1}^{C}}\sum\int_{W_m^{-1}(t_l)}^{W_m^{-1}(t_{l+1})}\lambda_c(s)ds}_{p_{l}^m}.
\end{aligned}
\end{eqnarray}
On the other hand, given current estimated parameters, we can calculate 
\begin{eqnarray}\label{qijm}
\begin{aligned}
q_{ij}^m&=\frac{\alpha_j\exp(-\beta_jW_m^{-1}(t_j^m))}{\sum_{j'}\alpha_{j'=1}^{J_i}\exp(-\beta_{j'}W_m^{-1}(t_{j'}^m))}\\
&=\frac{\alpha_j\exp(-\beta_jW_m^{-1}(t_j^m))}{\lambda_{c_i^m}(W_m^{-1}(t_i^m))},
\end{aligned}
\end{eqnarray}
and then apply Jensen's inequality to the term $\mathcal{B}$:
\begin{eqnarray}\label{termB}
\begin{aligned}
\mathcal{B}=&-\sideset{}{_{i=1}^{I_m}}\sum\log\Biggl(\sideset{}{_{j=1}^{J_i}}\sum\alpha_j\exp(-\beta_j W_m^{-1}(t_i^m))\Biggr)\\
\leq & \sideset{}{_{i=1}^{I_m}}\sum\sideset{}{_{j=1}^{J_i}}\sum q_{ij}^m\log\frac{q_{ij}^m}{\alpha_j\exp(-\beta_j W_m^{-1}(t_i^m))}\\
=&\sideset{}{_{i=1}^{I_m}}\sum\sideset{}{_{j=1}^{J_i}}\sum q_{ij}^m\Bigl( \log\frac{q_{ij}^m}{\alpha_j}+\beta_j W_m^{-1}(t_i^m)\Bigr)\\
=&\sum_{l=1}^{L-1}\sum_{t_i^m\in [t_l, t_{l+1})}\sum_{j=1}^{J_i} q_{ij}^m\beta_j (a_l^m t_i^m+b_l^m)))+\mathsf{C}\\
\end{aligned}
\end{eqnarray}

\subsection{Practical implementations}\label{ap3}
Taking a \textbf{multi-dimensional Hawkes process} as an example, we give the implementation details of our learning method.  
Specifically, the intensity function of the type-$c$ event at time $t$ is
\begin{eqnarray}
\begin{aligned}
\lambda_c(t)=\mu_c + \sum_{t_i<t}\phi_{c_i c_j}\exp(-w(t-t_i)),
\end{aligned}
\end{eqnarray}
where the parameter set $\theta$ consists of the background intensity vector $\bm{\mu}=[\mu_c]$ and the infectivity matrix $\bm{\Phi}=[\phi_{cc'}]$. 

\textbf{Maximum likelihood.} 
Given unwarped sequences $\{W_m^{-1}(S_m)\}_{m=1}^{M}$, we can maximize the likelihood of the sequences by an EM-based method~\cite{lewis2011nonparametric,zhou2013learning}. 
Specifically, the negative likelihood function and its tight upper bound can be written as
\begin{eqnarray*}
\begin{aligned}
&-\sum_{m=1}^{M}\log\mathcal{L}(\theta;W_m^{-1}(S_m))\\
=&\sum_{m=1}^{M}\Bigl[\sum_{c=1}^{C}\int_{0}^{T}\lambda_c(W_m^{-1}(s))ds\\
&-\sum_{i=1}^{I_m}\log\lambda_{c_i^m}(W_m^{-1}(t_i^m))\Bigr]\\
=&\sum_{m=1}^{M}\Bigl[\sum_{c=1}^{C}\Bigl(T\mu_c+\sum_{i=1}^{I_m}\phi_{cc_i^m}\int_{0}^{T-t_i^m}\exp(-wW_m^{-1}(s))ds\Bigr)\\
&-\sum_{i=1}^{I_m}\log\Bigl(\mu_{c_i^m}+\sum_{j=1}^{i-1}\phi_{c_i^mc_j^m}\exp(-w\tau_{ij})\Bigr)\Bigr]\\
\leq &\sum_{m=1}^{M}\Bigl[\sum_{c=1}^{C}\Bigl(T\mu_c+\sum_{i=1}^{I_m}\phi_{cc_i^m}\int_{0}^{T-t_i^m}\exp(-wW_m^{-1}(s))ds\Bigr)\\
&-\sum_{i=1}^{I_m}\Bigl(p_i\log\frac{\mu_{c_i^m}}{p_i}+\sum_{j=1}^{i-1}p_{ij}\log\frac{\phi_{c_i^mc_j^m}\exp(-w\tau_{ij})}{p_{ij}}\Bigr)\Bigr]\\
=&\mathcal{L}(\theta|\hat{\theta}).
\end{aligned}
\end{eqnarray*}
Here, $\tau_{ij}=W_m^{-1}(t_i^m)-W_m^{-1}(t_j^m)$ and $\hat{\theta}$ is current estimated parameters used to calculate $\{p_i,p_{ij}\}$ as
\begin{eqnarray}
\begin{aligned}
&p_i=\frac{\hat{\mu}}{\hat{\lambda}_{c_i^m}(W_m^{-1}(t_i^m))},\\ &p_{ij}=\frac{\hat{\phi}_{c_i^mc_j^m}\exp(-w\tau_{ij})}{\hat{\lambda}_{c_i^m}(W_m^{-1}(t_i^m))}.
\end{aligned}
\end{eqnarray}
As a result, we can update $\theta$ by minimizing $\mathcal{L}(\theta|\hat{\theta})$, which has the following closed-form solution:
\begin{eqnarray}\label{solution}
\begin{aligned}
&\mu_c = \frac{\sum_{m}\sum_{c_i^m=c}p_i}{MT},\\
&\phi_{cc'}=\frac{\sum_{m}\sum_{c_i^m=c}\sum_{c_j^m=c'}p_{ij}}{\sum_{m}\sum_{c_i^m=c'}\int_{0}^{T-t_i^m}\exp(-wW_m^{-1}(s))ds}.
\end{aligned}
\end{eqnarray}
According to the updated parameters, we can go back to calculate $\{p_i,p_{ij}\}$. 
Repeating the steps above till the objective function ($i.e.$, the negative log-likelihood) converges, we can obtain the optimum model given current $\{W_m\}_{m=1}^{M}$.

\textbf{Learning unwarping functions.} 
The key of this step is calculating the $\{p_{l}^m, q_{ij}^m\}$ mentioned in (\ref{termA}, \ref{termB}). 
For $p_{l}^m$, we have
\begin{eqnarray}\label{pcl}
\begin{aligned}
p_{l}^m=&\sum_{c=1}^{C}\int_{W_m^{-1}(t_l)}^{W_m^{-1}(t_{l+1})}\lambda_c(s)ds\\
=&\sum_{\substack{c=1,...,C\\t_i^m\in [t_l,t_{l+1})}}\Bigl(\phi_{cc_i^m}\int_{0}^{W_m^{-1}(t_{l+1})-W_m^{-1}(t_i^m)}e^{-ws}ds\\
&+\mu_c(W_m^{-1}(t_{l+1})-W_m^{-1}(t_{l}))\Bigr)\\
=&\sum_{\substack{c=1,...,C\\t_i^m\in [t_l,t_{l+1})}}\Bigl(\phi_{cc_i^m}\frac{1-e^{-wa_l^m(t_{l+1}-t_i^m)}}{w}\\
&+\mu_c a_l^m(t_{l+1}-t_{l})\Bigr).
\end{aligned}
\end{eqnarray}
For $q_{ij}^m$, because
\begin{eqnarray}
\begin{aligned}
&\lambda_{c_i^m}(W_m^{-1}(t_i^m))\\
=&\mu_{c_i^m} + \sum_{j=1}^{i-1}\phi_{c_i^mc_j^m}\exp(-w(W_m^{-1}(t_i^m)-W_m^{-1}(t_j^m)))\\
=&\sum_{j=0}^{i-1}\alpha_j\exp(-\beta_j W_m^{-1}(t_i^m)),
\end{aligned}
\end{eqnarray}
where for $j=0$, $\alpha_j=\mu_{c_i^m}$ and $\beta_j=0$; and for $j>0$, $\alpha_j = \phi_{c_i^mc_j^m}\exp(wW_m^{-1}(t_j^m))$ and $\beta_j=w$, 
we have
\begin{eqnarray}\label{qij}
\begin{aligned}
q_{ij}^m&=\frac{\alpha_j\exp(-\beta_jW_m^{-1}(t_i^m))}{\lambda_{c_i^m}(W_m^{-1}(t_i^m))}~\text{for}~j=0,...,i-1.
\end{aligned}
\end{eqnarray}

In our experiments, we configure our learning algorithm as follows. 
The number of landmarks $L=20$. 
The weight of regularizer $\gamma = 0.01$. 
The maximum number of outer iteration is $7$.
The maximum number of inner iteration for learning the Hawkes process model is $15$. 
The maximum number of inner iteration for updating warping functions is $5$. 
The interior-point method is applied. 

\subsection{The convexity of (\ref{opt:ab})}\label{ap4}
Ignoring constraints, (\ref{opt:ab}) can be decomposed into $2(L-1)$ problems with respect to each $a_l^m$ and $b_l^m$.
The objective function in (\ref{opt:ab}) that is related to $a_l^m$ can be formulated as 
\begin{eqnarray}
\begin{aligned}
f(x) = \frac{\alpha}{x}+\beta x + (x+\tau)^2,
\end{aligned}
\end{eqnarray}
where the unknown variable $x>0$, the coefficients $\alpha$ and $\beta$ are nonnegative, and $\tau$ is arbitrary. 
Because when $x>0$, $\frac{\alpha}{x}$, $\beta x$ and $(x+\tau)^2$ are convex functions, their sum, $i.e.$, $f(x)$, is also a convex function as well. 
Similarly, the objective function in (\ref{opt:ab}) that is related to $b_l^m$ can be formulated as 
\begin{eqnarray}
\begin{aligned}
f(x) = \beta x + (x+\tau)^2,
\end{aligned}
\end{eqnarray}
which is also a convex function. 

\begin{figure*}[t]
\centering
\includegraphics[width=0.95\linewidth]{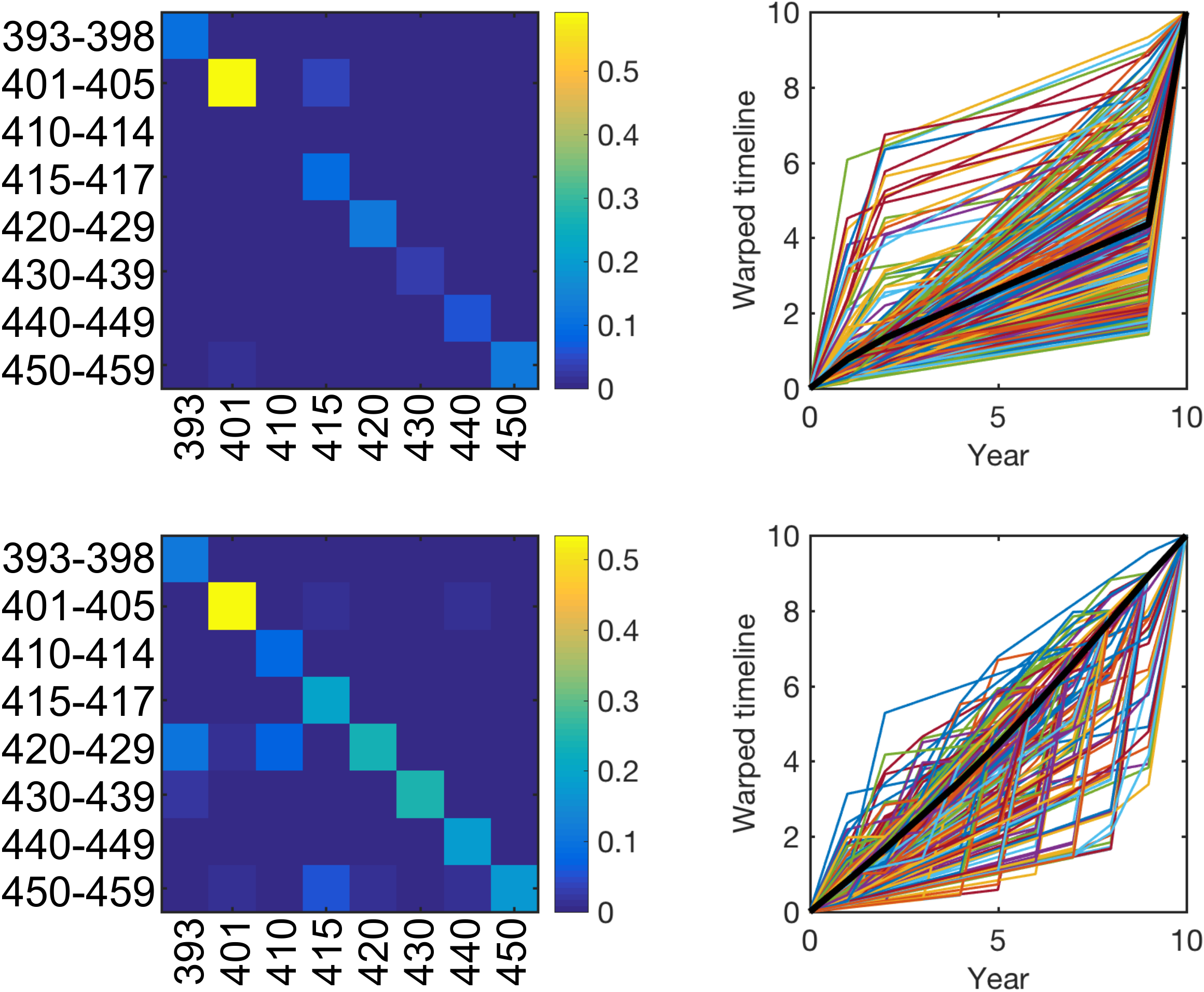}
\caption{\small Experimental result of WLR (top) and our method (bottom) on MIMIC III data.\label{real_mimic}}
\end{figure*}

\begin{figure*}[t]
\centering
\includegraphics[width=0.95\linewidth]{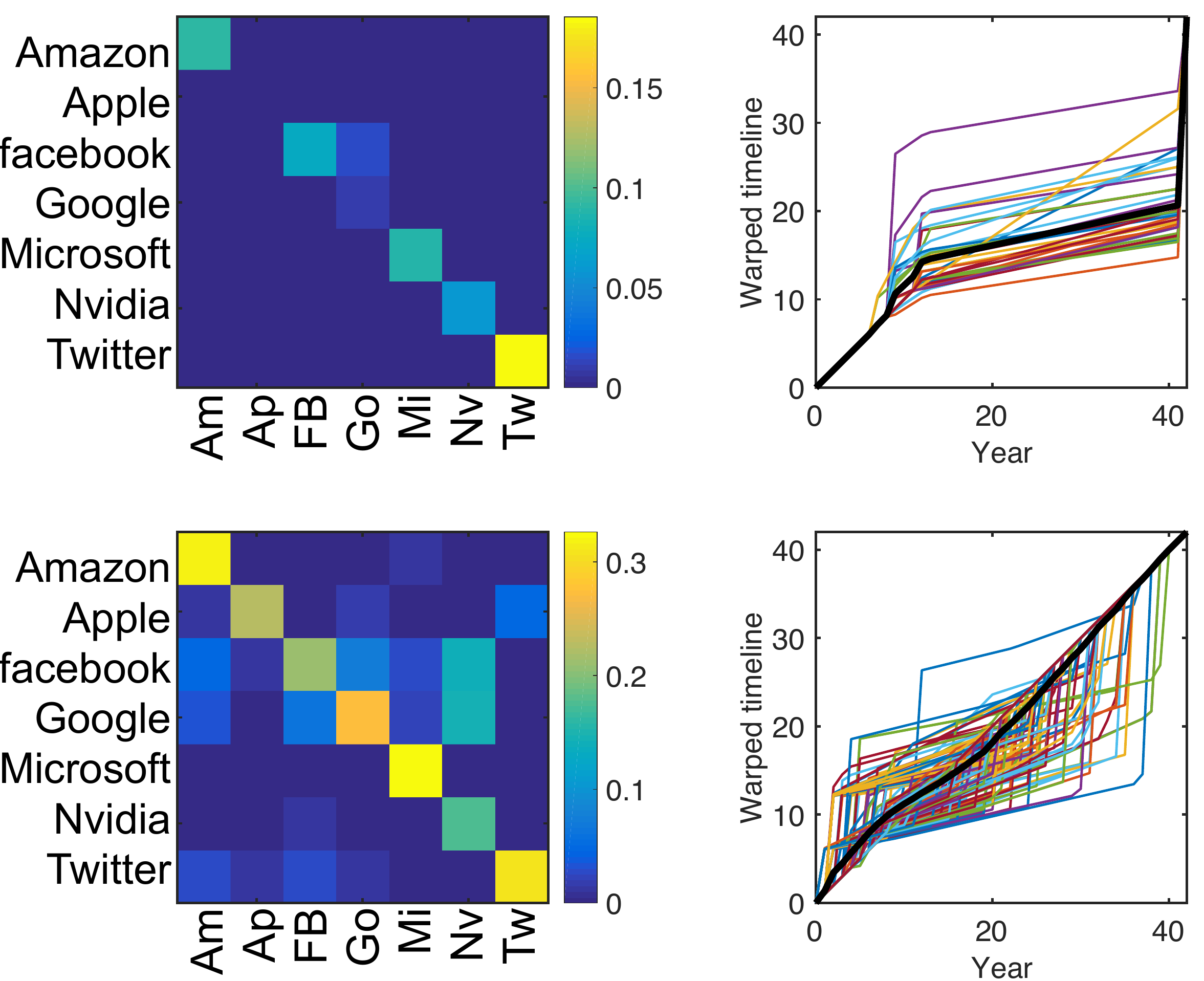}
\caption{\small Experimental result of WLR (top) and our method (bottom) on LinkedIn data.\label{real_linkedin}}
\end{figure*}

\subsection{Generating warping/unwarping functions}\label{ap5}
For the synthetic data used in our experiments, each warping function in $[0,T]$ is represented by a set of local cosine basis as
\begin{eqnarray}
\begin{aligned}
W_m(t) &= \sideset{}{_{n=1}^{N}}\sum w_n^m f_n(t),\\
f_n(t) &= \begin{cases}
\cos^2(\frac{\pi}{2\Delta}(t - t_n)), & |t-t_n|<=\Delta\\
0, & \text{otherwise}.
\end{cases}
\end{aligned}
\end{eqnarray}
The time window $[0,T]$ is segmented by $N$ landmarks $\{t_n\}_{n=1}^{N}$, where $t_1=0$ and $t_N=T$. 
For each $f_n(t)$, the landmark $t_n$ is its center and $\Delta$ is the distance between adjacent landmarks. 
The first $N-1$ coefficients $\{w_n^m\}_{n=1}^{N-1}$ is sampled from $[0,T]$ uniformly and sorted by ascending order. 
The last coefficient $w_N^m$ is set to be $T$. 
Using this method, we can ensure that all warping functions (and the corresponding unwarping functions) are monotone increasing maps from $[0,T]$ to $[0,T]$ and their average is close to an identity function.

\subsection{Details of experiments}\label{ap6}
For the MIMIC data set, each admission is associated with a set of diagnose. 
Based on the priority assigned to the diagnose, we only keep the ICD-9 code with the highest priority as the event type of the admission. 
In our work, we assume that the admission behaviors of all patients happen from 2001 to 2012 or their death date.
In this case, the length of time window $T$ is different for each patient. 
Our learning method can be extended to adjust this situation. 
In particular, we can use specific $T$'s for different event sequences, $i.e.$, replacing $T$ to $T^m$ in our model and learning algorithm. 
For our piecewise linear model, the distance between adjacent landmarks can be adjusted as well according to $T^m$. 
For each patient in the MIMIC data set, we can set the time stamp of its last admission event as $T^m$.

The categories of the diseases of circulatory system are shown below:
\begin{enumerate}
\item Chronic rheumatic heart disease (ICD-9: 393 - 398)
\item Hypertensive disease (ICD-9: 401 - 405)
\item Ischemic heart disease (ICD-9: 410 - 414)
\item Diseases of pulmonary circulation (ICD-9: 415 - 417)
\item Other forms of heart disease (ICD-9: 420 - 429)
\item Cerebrovascular disease (ICD-9: 430 - 438)
\item Diseases of arteries, arterioles, and capillaries (ICD-9: 440 - 449)
\item Diseases of veins and lymphatics, and other diseases of circulatory system (451 - 459)
\end{enumerate}

Using our RPP method, we learn a $8$-dimensional Hawkes process from $1,129$ patient's admission records.
Compared to synthetic data, the MIMIC III dataset is sparse ($i.e.$, most of the patients have just $2$ - $5$ admission events), so we use a larger weight for regularizer ($i.e.$, $\gamma = 10$) and fewer landmarks ($i.e.$, $L=5$).
Similarly, we can learn a $7$-dimensional Hawkes process from $709$ users' job hopping records, in which we also set $\gamma=10$ and $L=5$.

These infectivity matrices further verify the justifiability of our learning method because they reflect some reasonable phenomena.
In Fig.~\ref{real_mimic}, we can find that: 
\begin{enumerate}
\item All disease categories have strong self-triggering patterns. 
The ``hypertension disease'' (ICD-9 code 404-405), which is one of the most common disease in modern society, has the strongest self-triggering pattern --- the value of the second diagonal element is over $0.5$
It means that for a patient suffering to a certain disease of circulatory system, he or she is likely to re-admit to hospital in next 10 years for the same disease. 
\item The $5$-th row in Fig.~\ref{real_mimic} corresponds to the category ``other forms of heart disease'' (ICD-9 code 420-429).
According to its name we can know that this category contains many miscellaneous heart diseases and should have complicated relationships with other categories. 
Our learning result reflects this fact --- the $5$-th row of our infectivity matrix contains many non-zero elements, which means that this disease category can be triggered by other disease categories.
\end{enumerate}

In Fig.~\ref{real_linkedin}, we can find that: 
\begin{enumerate}
\item All IT companies have strong self-triggering patterns, which means that most of employees are satisfied to their companies. 
Especially for Amazon and Microsoft, their diagonal elements are over $0.3$. 
It means that the expected happening rate of internal promotion for their employees is about $0.3$ event per year.
\item The elements of ``Facebook-Google'' and ``Google-Facebook'' pairs are with high values, which means that job hopping happens frequently between Facebook and Google. 
This result reflects their fierce competition. 
\item The elements of ``Facebook-Nvidia'' and ``Google-Nvidia'' are with high values, which reflects the fact that recent years many Nvidia's employees jump to Google and Facebook to develop hardware and systems of AI.  
\end{enumerate}

In our opinion, there are three reasons for the increased performance. 
Firstly, our parametric model is more robust to data insufficiency, which can capture complicated mechanism of event sequences from relatively fewer observations. 
Secondly, we learn the registered model and the warping functions in an alterative, rather than independent way, to avoid serious model misspecification, and such a method has a good convergence. 
Thirdly, the proposed piecewise linear model has a good capability to describe warping function approximately, which achieves a trade-off between the complexity of the model and the performance.

\end{document}